\pdfoutput=1

\documentclass[11pt]{article}

\usepackage[]{acl}

\usepackage{times}
\usepackage{latexsym}

\usepackage[T1]{fontenc}

\usepackage[utf8]{inputenc}

\usepackage{microtype}

\usepackage{CJKutf8}

\newcommand{\met}{MeTNet}
\newcommand{\fcm}{FEW-COMM}

\usepackage{bbm}
\usepackage{pifont}

\usepackage{subfig}
\usepackage{multirow}
\usepackage{arydshln}

\usepackage{amsmath, amsfonts}
\usepackage{graphicx}

\usepackage{algorithmicx}
\usepackage{algorithm}
\usepackage{algpseudocode}

\makeatletter
\newcommand{\removelatexerror}{\let\@latex@error\@gobble}
\makeatother

\algnewcommand{\LeftComment}[1]{\State \(\triangleright\) #1}

\usepackage{xcolor}

\title{Meta-Learning Triplet Network with Adaptive Margins for Few-Shot Named Entity Recognition}

\author{Chengcheng Han\textsuperscript{1\thanks{\ \ Equal contribution}}\ \ \ \ Renyu Zhu\textsuperscript{1\footnotemark[1]}\ \ \ \ Jun Kuang\textsuperscript{2}\ \ \ \ FengJiao Chen\textsuperscript{2} \\
\bf{Xiang Li\textsuperscript{1\thanks{\ \ Corresponding author}}}\ \ \ \ \bf{Ming Gao\textsuperscript{1}} \ \ \ \ \bf{Xuezhi Cao\textsuperscript{2}}\ \ \ \ \bf{Wei Wu\textsuperscript{2}} \\
    \textsuperscript{1}School of Data Science and Engineering, East China Normal University, Shanghai, China \\  \textsuperscript{2}Meituan Inc., Beijing, China \\ 
    \texttt{\{52215903007,52175100003\}@stu.ecnu.edu.cn}  
    \\  \texttt{\{xiangli,mgao\}@dase.ecnu.edu.cn}
    \\  \texttt{\{kuangjun,chenfengjiao02,caoxuezhi,wuwei30\}@meituan.com} }

\begin{document}
\maketitle
\begin{abstract}
Meta-learning methods have been widely used in few-shot named entity recognition~(NER), especially prototype-based methods.
However,
the \texttt{Other(O)}
class is difficult to be represented by a prototype vector
because 
there are generally a large number of 
samples in the class 
that have miscellaneous semantics.
To solve the problem,
we 
propose \met, 
which generates prototype vectors for entity types only but not \texttt{O}-class.
We design an improved triplet network 
to map samples and prototype vectors into a low-dimensional space
that is easier to be classified
and
propose an adaptive margin for each entity type.
The margin plays as a radius 
and controls a region with adaptive size
in the low-dimensional space.
Based on the regions, we propose a new inference procedure to predict the label of a query instance.
We conduct extensive experiments in both 
in-domain 
and cross-domain settings to show the superiority of \met\ over other 
state-of-the-art methods.
In particular,
we release a Chinese few-shot NER dataset \fcm\ extracted from a well-known e-commerce platform.
To the best of our knowledge, this is the first Chinese few-shot NER dataset.
All the datasets and codes are provided at \url{https://github.com/hccngu/MeTNet}.
\end{abstract}

\section{Introduction}
\label{intro}

\begin{figure}[!t]
    \centering
    \subfloat[]{
    \label{fig:example}
    \includegraphics[width=0.45\textwidth]{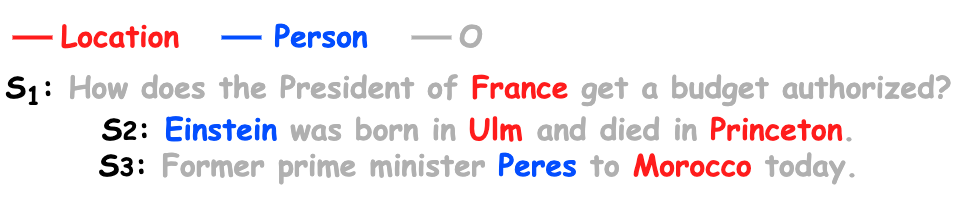}
    }
    \\
    \subfloat[]{
    \label{fig:vector}
    \includegraphics[width=0.45\textwidth]{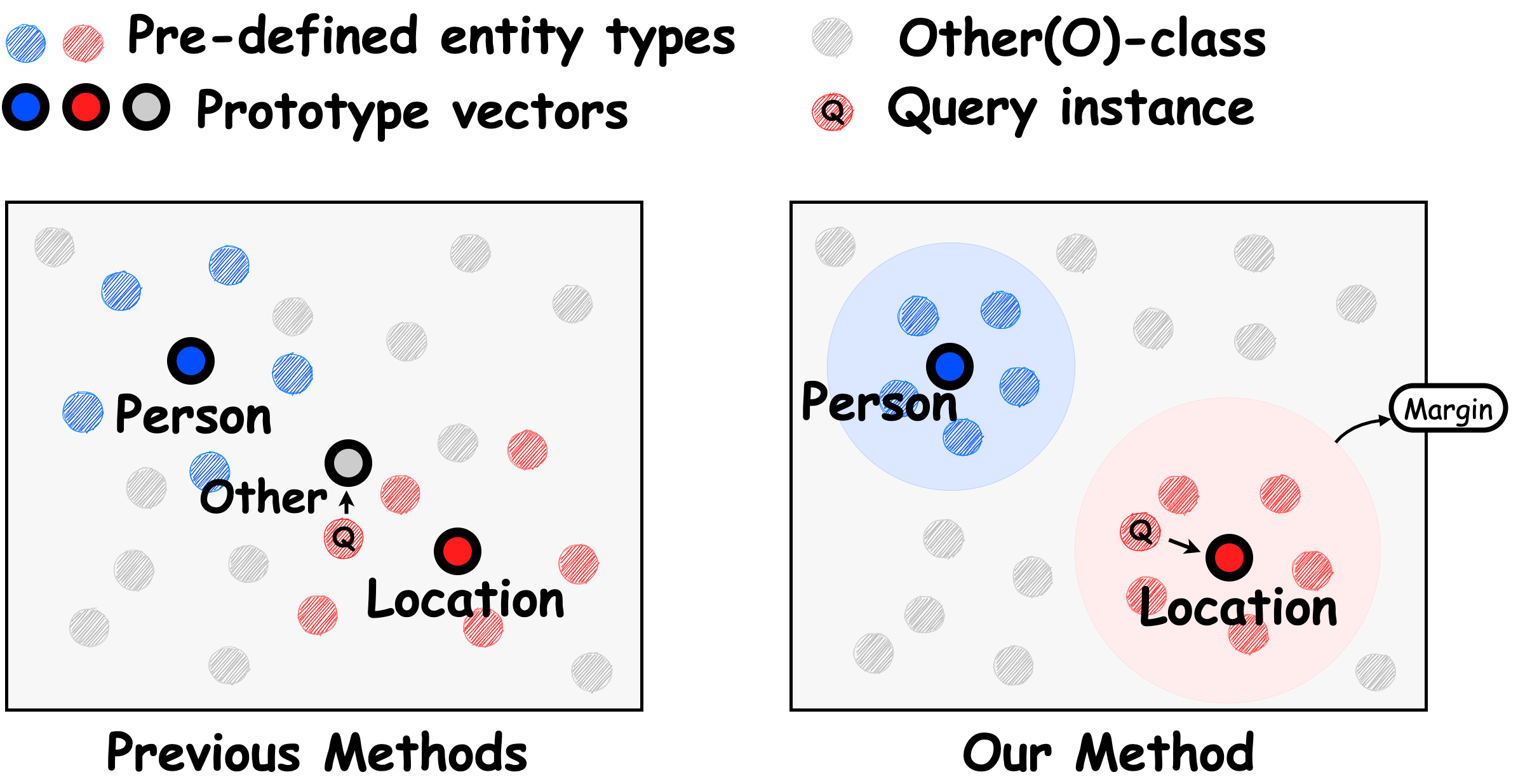}
    }
    \caption{(a): Samples in \texttt{O}-class are semantically different. (b): The comparison between previous methods and ours to handle \texttt{O}-class. 
    \textbf{Left:} 
    Since the 
    query instance whose true label is \texttt{Location} is closest to the 
    prototype vector of \texttt{O}-class, previous methods misclassify it to \texttt{O}-class.
    \textbf{Right:}
    We compute prototype vectors for entity types only and 
    learn an adaptive margin for each entity type to determine a region.
    Samples in the region of a class are labeled with the class,
    while
    samples outside of all the regions are predicted to be in \texttt{O}-class.}
    \label{fig1}
\end{figure}

Named entity recognition~(NER), 
as a fundamental task in information extraction~\cite{IE:conf/kdd/RitterMEC12},
aims to locate and classify words or expressions into
\emph{pre-defined entity types}, 
such as \texttt{persons}, \texttt{organizations}, \texttt{locations}, \texttt{dates} and \texttt{quantities}.
While
a considerable number of approaches based on deep neural networks have shown remarkable success in NER, 
they generally require massive labeled data as training set.
Unfortunately, 
in some specific domains, named entities that need professional knowledge to understand are difficult to be manually annotated in a large scale.

To address the problem, 
few-shot NER has been studied, 
which aims to recognize unseen entity types with few annotations.
In particular,
some models~\cite{Proto19:conf/sac/FritzlerLK19,TapNetCDT:conf/acl/HouCLZLLL20,ESD:journals/corr/abs-2109-13023} are proposed 
based on the prototypical network~(PROTO)~\cite{PROTO:conf/nips/SnellSZ17}, 
which is
a popular meta-learning method. 
The general procedure of these prototype-based NER models is summarized as follows.
First,
they generate a prototype vector for each class, including both entity types and \texttt{Other(O)} class, 
to represent the class.
Then they compute 
the distance between a query sample~(instance)~\footnote{We interchangeably use sample and instance in this paper.}
and all these prototype vectors, and predict the query instance to the class with the smallest distance.
However,
for NER,
the \texttt{O}-class covers all the miscellaneous words that are not classified as entity types.
These words could span a wide range of semantics.
For example, 
in Figure~\ref{fig:example},
the words ``was'', ``president'', ``budget'' and ``today'' are semantically different even if they all belong to \texttt{O}-class.
A single prototype vector would thus be insufficient to model the miscellaneous semantics of \texttt{O}-class,
which could further lead to the incorrect prediction of query instances~(see Figure~\ref{fig:vector}).

In this paper,
to solve the issue,
we propose to generate prototype vectors only for entity types but not \texttt{O}-class.
In particular,
we design 
a 
\textbf{Me}ta-Learning \textbf{T}riplet \textbf{Net}work with adaptive margins, namely, \met,
to map samples and prototype vectors into a low-dimensional space,
where the inter-class distance between samples is enlarged
and the intra-class distance between samples and their corresponding prototype vectors is shortened.
We further 
design an improved triplet loss function with adaptive margins, 
which assigns different weights to samples,
minimizes the absolute distance between an anchor and a positive sample,
and maximizes the absolute distance between an anchor and a negative sample.
The adaptive margin 
plays as a radius and controls a region
for each entity type
in the low-dimensional space~(see Figure~\ref{fig:vector}).
Based on
these regions,
we further propose a novel inference procedure.
Specifically,
given a query instance,
we predict it to be in \texttt{O}-class,
if it is located outside all the regions;
otherwise,
we label it with the entity type of its located region.
Further,
if it is contained in multiple regions,
we label it with the entity type that has the smallest distance between the query instance and the region center.
Finally, 
we summarize our main contributions in this paper as follows.

\begin{itemize}
    \item We propose an improved triplet network with adaptive margins~(\met)
    and a new inference procedure for few-shot NER.

    
    \item We release the first Chinese few-shot NER dataset \fcm, to our best knowledge.
    
    \item 
    We perform extensive experiments to show the superiority of \met\ over other competitors.
    
\end{itemize}

\section{Related Work}

\subsection{Meta-Learning}

Meta-learning,
also known as ``learning to learn'', 
aims to train models to adapt to new tasks rapidly with few training samples.
Some
existing methods~\cite{PROTO:conf/nips/SnellSZ17, vinyals2016matching_network}
are based on metric learning. 
For example,
Matching Network~\cite{vinyals2016matching_network} 
computes similarities between support sets and query instances,
while
the prototypical network~\cite{PROTO:conf/nips/SnellSZ17} learns a prototype vector for each class and classifies query instances based on the nearest prototype vector.
Other representative metric-based methods include 
Siamese Network~\cite{SN_koch2015siamese}
and 
Relation Network~\cite{Relation_network_sung2018learning}.
Further, 
some approaches,
such as MAML~\cite{MAML_finn2017model}
and Reptile~\cite{Reptile_nichol2018first}, 
are optimization-based,
which aim to train a meta-learner as an optimizer or adjust the optimization process.
There also exist model-based methods,
which
learn a hidden feature space
and predict the label of a query instance in an end-to-end manner.
Compared with the optimization-based methods, 
model-based methods could be easier to optimize but less generalizable to out-of-distribution tasks~\cite{metaSurvey:journals/corr/abs-2004-05439}.
The representative 
model-based 
methods include 
MANNs~\cite{MANN_santoro2016meta}, 
Meta networks~\cite{Meta_network_munkhdalai2017meta}, SNAIL~\cite{SNAIL_mishra2017simple} and CPN~\cite{CNP_garnelo2018conditional}.

\subsection{Few-shot NER}
Few-shot NER has recently received great attention~\cite{Illi_huang2021few,Container_das2021container,DecomposedMETA_ma2022decomposed}
and
meta-learning-based methods have been applied to solve the problem.
For example,
\citet{Proto19:conf/sac/FritzlerLK19} combine PROTO~\cite{PROTO:conf/nips/SnellSZ17} with conditional random field for few-shot NER.
Inspired by the nearest neighbor inference~\cite{nn_wiseman2019label},
StructShot~\cite{StructShot_yang2020simple} employs structured nearest neighbor learning and Viterbi algorithm to further improve PROTO.
MUCO~\cite{MUCO_tong2021learning} trains a binary classifier 
to learn multiple prototype vectors for representing miscellaneous semantics of \texttt{O}-class.
ESD~\cite{ESD:journals/corr/abs-2109-13023} uses various types of attention based on PROTO to improve the model performance.
However,
most of these methods use one or multiple prototype vectors to represent \texttt{O}-class, 
while we 
compute 
prototype vectors for entity types only
and further
design a new inference procedure.

Very recently, prompt-based techniques have also been applied in few-shot NER~\cite{BARTNER_cui2021template,FudanPrompt_ma2021template,LightPrompt_chen2021lightner,ProtoPrompt_cui2022prototypical}.
However,
the performance
of these methods is very unstable, 
which heavily
depend on the designed prompts~\cite{BARTNER_cui2021template}.
Thus,
without a large validation set,
their applicability is limited in few-shot learning.


\section{Background}

\subsection{Problem Definition}
\label{sec:PD}
A training set $\mathcal{D}_{train}$ 
consists of word sequences and their label sequences.
Given a word sequence
$X = \{x_1, ..., x_n\}$,
we denote
$L = \{l_1, ..., l_n\}$ as its corresponding label sequence.
We use 
$\mathcal{Y}_{train}$ to denote the label set of the training data
and 
$l_i \in \mathcal{Y}_{train}$.
In addition,
given a test set $\mathcal{D}_{test}$, 
let $\mathcal{Y}_{test}$ denote the label set of the test set, which 
satisfies $\mathcal{Y}_{train} \cap \mathcal{Y}_{test} = \emptyset$.
Our goal is to develop a model that learns from $\mathcal{D}_{train}$
and then makes predictions for unseen classes in $\mathcal{Y}_{test}$, for which we only have few annotations.

\subsection{Meta-training}
Meta-learning methods include two stages: meta-training and meta-testing.
In meta-training, the model is trained on meta-tasks sampled from $\mathcal{D}_{train}$.
Each meta-task contains a support set and a query set.
To create a training meta-task, 
we first sample $N$ classes from $\mathcal{Y}_{train}$.
After that,
for each of these $N$ classes, 
we sample $K$ instances as the support set $\mathcal{S}$ and $L$ instances as the query set $\mathcal{Q}$.
The support set is similar as the training set in the traditional supervised learning 
but it only contains a few samples;
the query set acts as the test set but it can be used to compute gradients for updating model parameters in meta-training stage.
Given the support set, we refer to the task of making predictions over the query set as \emph{$N$-way $K$-shot classification}.




\subsection{Meta-testing}
In the testing stage, 
we also use meta-tasks to test whether our model can adapt quickly to new classes. 
To create a testing meta-task, 
we first sample $N$ new classes from $\mathcal{Y}_{test}$. 
Similar as in meta-training,
we then sample the support set and the query set from the $N$ classes, respectively.
The support set is used for fine-tuning while the query set is for testing.
Finally,
we evaluate the average performance on the query sets across all testing meta-tasks.

\section{Method}
\label{sec:method}

\begin{figure*}[htbp]
    \centering
    \includegraphics[height=200pt, width=410pt]{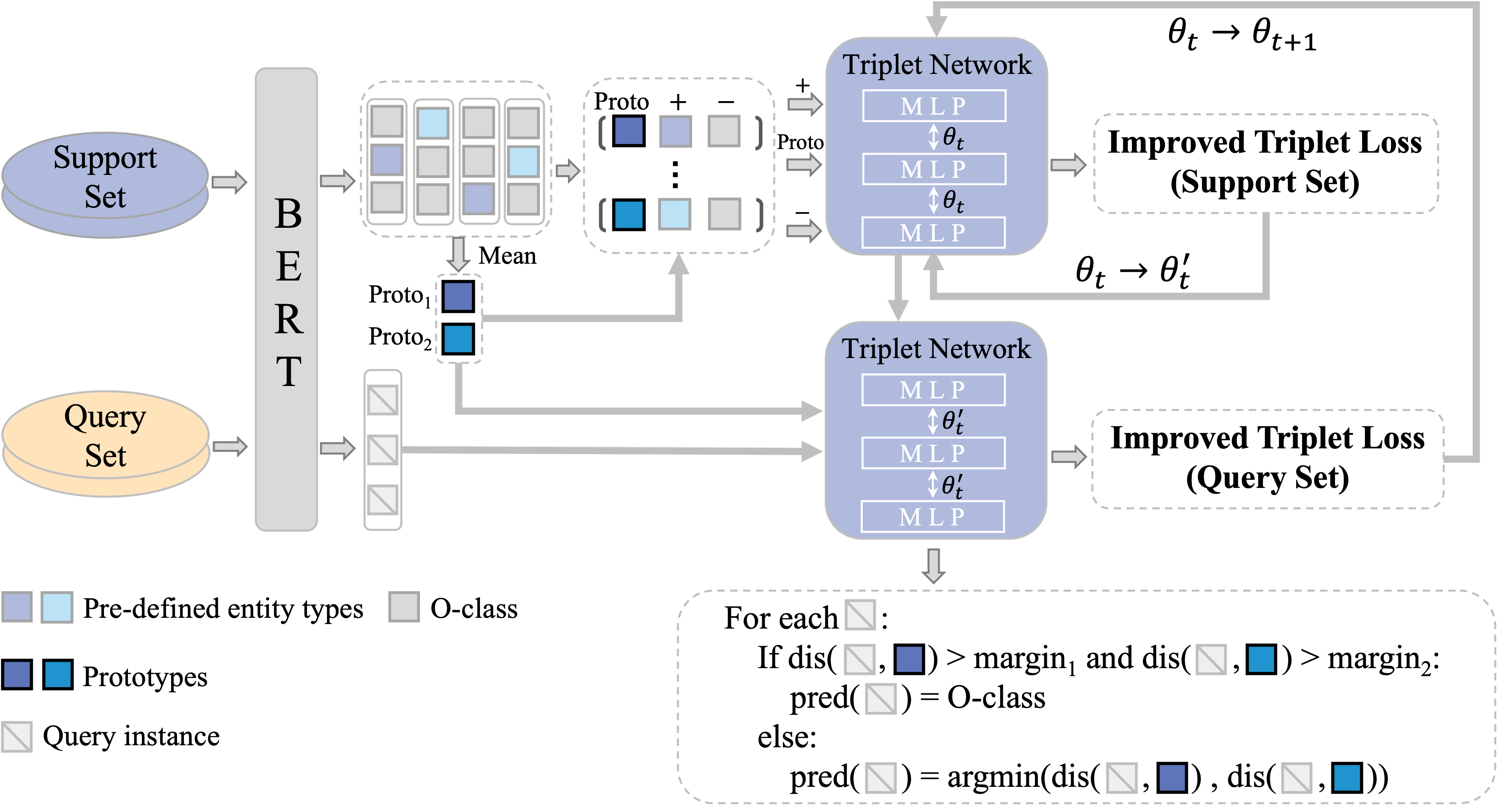}
    \caption{The overall architecture of \met\ for a $2$-way $2$-shot problem.}
    \label{fig:model}
\end{figure*}

In this section, we describe our \met\ algorithm.
We first give an overview of \met, which is illustrated in Figure~\ref{fig:model}.
\met\ first represents samples with BERT text encoder,
based on which the embeddings of words and prototype vectors are initialized.
Then it generates triples based on the support sets and prototype vectors, 
and employs an improved triplet network with adaptive margins to map
words and prototype vectors into a space that is much easier to classify.
For each entity type,
an adaptive margin plays as a radius 
and controls a region centered at the corresponding prototype vector.
These regions are further used in the inference stage.
Next, we describe each component of \met\ in detail.

\subsection{Text Encoder}
\label{model:text_encoder}

We first represent each word in a low-dimensional embedding vector. 
Following~\cite{StructShot_yang2020simple,fewnerd_ding2021few},
we use BERT~\cite{bert_devlin2018bert} as our text encoder.
Specifically,
given a sequence of $n$ words~$[x_1, x_2, ..., x_n]$,
we take the output of the
final hidden layer in BERT as the initial representations $\mathbf{h}_i$ for $x_i$:  
\begin{eqnarray}
\label{eq:bert}
    [\mathbf{h}_1, \mathbf{h}_2, ..., \mathbf{h}_n] = \texttt{BERT}_{\phi}([x_1, x_2, ..., x_n]),
\end{eqnarray}
where $\phi$ represents parameters of BERT.
Then for each pre-defined entity type~$c_j$, 
we construct its initial prototype vector $\mathbf{h}_{c_j}$ by averaging the representations of words labeled as $c_j$.

\subsection{Triplet Network}
\label{model:triplet_network}

A triplet network~\cite{tripletNetwork_hoffer2015deep} 
is composed of three sub-networks,
which have the same network architecture 
with shared parameters to be learned.
For the triplet network,
triples are taken as its inputs. 
Each triple consists of an anchor, a positive sample and a negative sample,
and 
we feed each sample into 
a sub-network.

\paragraph{Construct Triples}
We first construct triples for different entity types.
Specifically,
for each entity type,
we take its prototype vector as the anchor,
instances in the entity type as positive samples,
and other instances as negative ones.
Since the number of negative samples is generally larger than that of positive samples,
we select $k$ negative samples with 
the nearest distance to
the prototype vector.
After that,
for each positive sample and each negative sample,
we construct triples, respectively.


\paragraph{Improved Triplet Loss}
Given 
the distance~$d_p$ between the anchor and the positive sample,
and
the distance~$d_n$ between the anchor and the negative sample,
the original triplet loss aims
to optimize the \emph{relative distance} among the anchor, the positive sample and the negative sample, which is formulated as:
\begin{eqnarray}
\label{eq:TL}
    \mathcal{L}_{T} &=& max(0, m+d_p-d_n), \\
    d_p &=& d(f_{\theta}(\mathbf{h}_a), f_{\theta}(\mathbf{h}_p)), \\
    d_n &=& d(f_{\theta}(\mathbf{h}_a), f_{\theta}(\mathbf{h}_n)),
\end{eqnarray}
where $m$ is a margin, 
$d(\cdot, \cdot)$ denotes the Euclidean distance function
and
$f_{\theta}(\cdot)$ is the embedding vector
generated from the triplet network.
However,
there exist three main problems in
the original triplet loss function.
First,
the original triplet loss pays more attention to the relative distance between $d_p$ and $d_n$.
When $d_p$ and $d_n$ are both large
but their difference is small, the loss will be small.
But our goal is to optimize absolute size of $d_p$ and $d_n$.
Second,
the loss function considers all the samples are equally important, but  
their importance is empirically relevant to their distance to the anchor.
Third,
the margin is fixed and unique.
However, 
different entity types generally correspond to regions with various sizes.
To address these problems, 
we design an improved triplet loss as follows:
\begin{eqnarray}
\label{eq:improvedTL}
\hspace{-2mm}
    \mathcal{L}_{IT} \!\!\!\!&=&\!\!\!\! \frac{\alpha}{1+e^{-(d_p-m_i)}} \cdot d_p \nonumber\\
    \!\!\!\!&+&\!\!\!\! \frac{1-\alpha}{1+e^{-(m_i-d_n)}} \cdot max(m_i-d_n, 0),
\end{eqnarray}
where $\alpha$ is a balancing weight
and $m_i$ denotes a learnable margin of entity type $c_i$.
In Equation~\ref{eq:improvedTL},
we separately optimize the \emph{absolute distances} $d_p$ 
and $d_n$.
On the one hand,
we directly minimize $d_p$.
On the other hand,
considering that each entity type uses a region to include positive samples,
we thus maximize $d_n$ by pushing the negative sample away from the region.
Further,
we assign different weights to samples based on their distances to anchors.
Intuitively, 
the farther the positive samples  
or the closer the negative samples are to the anchors,
the larger the weights should be given to amplify the loss.
Finally,
we set adaptive margins for different entity types,
which play as region radiuses and control region sizes.


\subsection{Inference}
\label{sec:inference}

\begin{figure}[!t]
    \centering
    \includegraphics[width=0.45\textwidth]{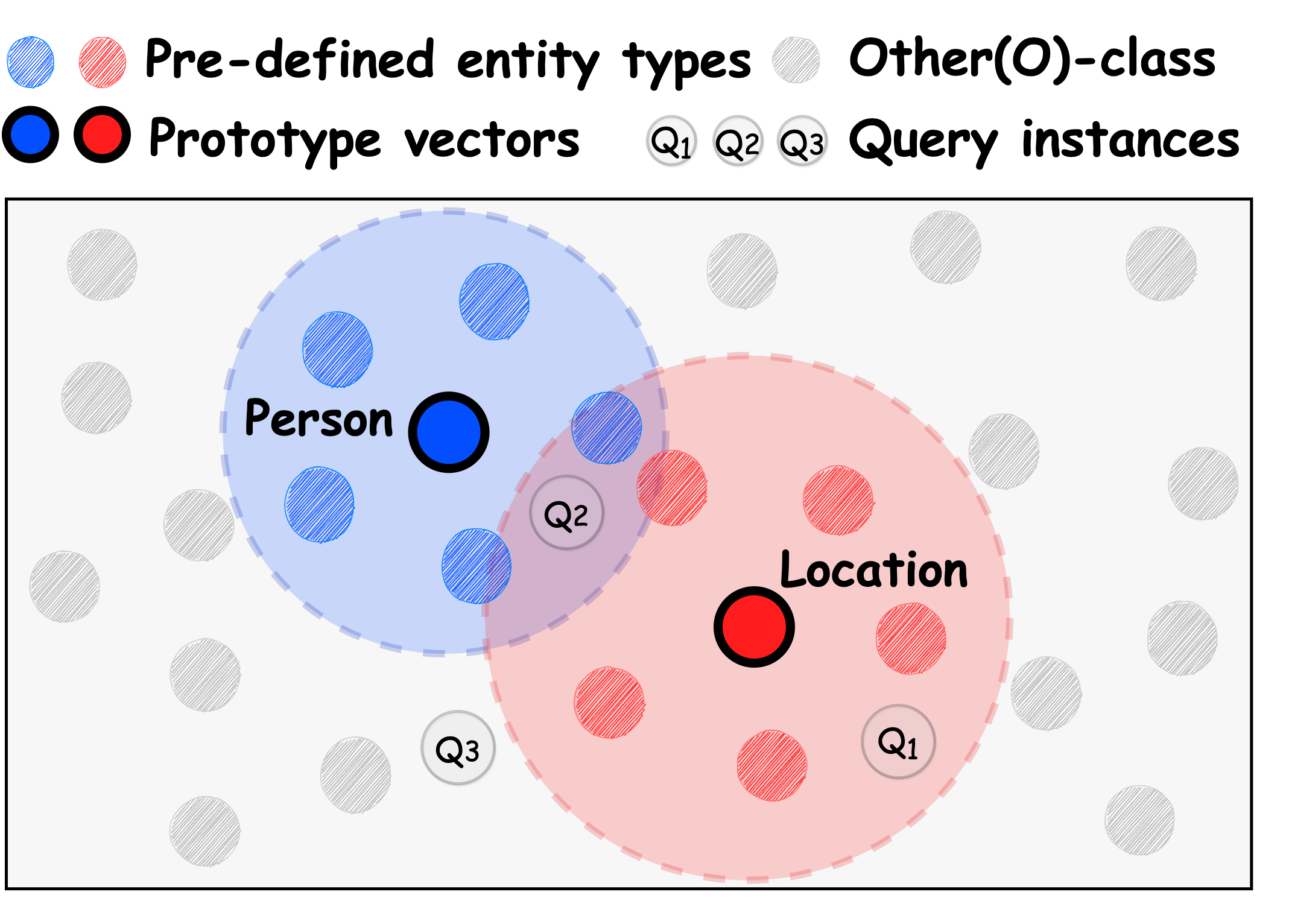}
    \caption{An example to illustrate the inference procedure in \met. 
    The dashed circles represent the regions of pre-defined entity types determined by adaptive margins.
    The labels of 
    $Q_1$, $Q_2$ and $Q_3$ are predicted to be \texttt{Location},
    \texttt{Person} and \texttt{O}-class, respectively.}
    \label{fig:inference}
\end{figure}

In the inference stage,
most existing methods 
calculate the distances between a query instance
and all the prototype vectors for both entity types and \texttt{O}-class,
and predict the query instance to be in the class with the smallest distance.
Different from these methods,
our model avoids handling \texttt{O}-class directly.
Instead,
we make predictions based on the regions of entity types.
As shown in Figure~\ref{fig:inference},
the entity types \texttt{Person} and \texttt{Location}
have their own regions 
controlled by different margins.
When a query instance (e.g., $Q_1$)
is only located in one region,
we label it with the entity type corresponding to the located region;
when a query instance (e.g., $Q_2$) is contained in multiple regions, 
we calculate its distances to different region centers and predict its entity type to be that with the smallest distance;
when a query instance (e.g., $Q_3$) is outside all the regions,
it is labeled with \texttt{O}-class.

\subsection{Training Procedure}
\label{model:training_procedure}
Inspired by MAML~\cite{MAML_finn2017model},
we first update the model parameters~$\theta$ with samples in the support set:
\begin{eqnarray}
\label{eq:maml_support}
\theta' = \theta - \gamma \nabla_{\theta}\mathcal{L}_{IT}(\theta; \mathcal{S}),
\end{eqnarray}
where $\gamma$ is the learning rate and $\mathcal{S}$ represents the support set. 
With few-step updates, 
$\theta$ becomes $\theta'$.
Then
based on $\theta'$, the triplet network can map query instances and prototype vectors into a low-dimensional space that is much easier to classify.
After that, we update the model parameters~$\theta$ with samples in the query set:
\begin{eqnarray}
\label{eq:maml_query}
\theta \leftarrow \theta - \beta \nabla_{\theta}\mathcal{L}_{IT}(\theta'; \mathcal{Q}),
\end{eqnarray}
where $\beta$ is the meta learning rate and $\mathcal{Q}$ represents the query set.
This optimization simulates the testing process in the training stage and boosts the generalizability of the model to unseen classes with only few-step updates.
The overall procedure of \met\ is summarized in  Appendix~\ref{app:alg_MeTNet}.

\section{Experiments}
In this section, 
we comprehensively evaluate the performance of \met\
in both
in-domain and cross-domain settings. 
The in-domain setting indicates that both the training set and the test set come from the same domain,
while the cross-domain setting indicates that they are from different domains.

\subsection{Datesets}
\label{sec:datasets}

We use four public English datasets
and one new Chinese dataset.
Statistics of these datasets are given in Appendix~\ref{app:datasets}.
For the English datasets,
they are
FEW-NERD~\cite{fewnerd_ding2021few},
WNUT17~\cite{wnut_derczynski2017results},
Restaurant~\cite{Restaurant_liu2013asgard} and 
Multiwoz~\cite{multiwoz_budzianowski2018multiwoz}.
Specifically,
FEW-NERD designs an annotation schema of 8 coarse-grained (e.g., ``Person'') entity types and 66 fine-grained (e.g., ``Person-Artist'') entity types, and constructs two tasks. 
One is FEW-NERD-INTRA,
where all the entities in the training set (source domain), validation set and test set (target domain) belong to different coarse-grained types.
The other is FEW-NERD-INTER, 
where only the fine-grained entity types are mutually disjoint in different sets.
We conduct in-domain experiments on both tasks.
To further validate the model's generalizability on cross-domain tasks, 
we also use three NER datasets from different domains, namely WNUT17 (Social),
Restaurant (Review) and
Multiwoz (Dialogue).

We also construct and conduct experiments on
a Chinese few-shot NER dataset, namely,
\fcm.
The dataset consists of 66,165 product description texts that merchants display on a large e-commerce platform, including 140,936 entities and 92 pre-defined entity types.
These entity types are various commodity attributes that are manually defined by domain experts, such as ``material'', ``color'' and ``origin''.
Specifically,
we first hire five well-trained annotators to label the texts in one month
and then ask four domain experts to review and rectify the results.
To the best of our knowledge, 
it is the first Chinese dataset specially constructed for few-shot NER.
Due to the space limitation, please see Appendix~\ref{appendix:comm} for more details on the dataset.

\begin{table*}[htbp]
	\begin{center}
	\resizebox{2.0\columnwidth}{!}{
		\begin{tabular}{cccccccccc}
			\hline
			\hline
			\multirow{2}{*}{Method}&\multicolumn{4}{c}{FEW-NERD-INTER}&\multicolumn{4}{c}{FEW-NERD-INTRA}&\multirow{2}{*}{Average} \\
			\cline{2-9}
            &$5$-$1$&$5$-$5$&$10$-$1$&$10$-$5$&$5$-$1$&$5$-$5$&$10$-$1$&$10$-$5$& \\
            \hline
			MAML  
			& 38.52\scriptsize $\pm$0.67 & 49.86\scriptsize $\pm$0.33 & 30.20\scriptsize $\pm$0.78 & 33.39\scriptsize $\pm$0.49 & 30.14\scriptsize $\pm$0.53 & 38.38\scriptsize $\pm$0.41 & 23.05\scriptsize $\pm$0.45 & 28.52\scriptsize $\pm$0.59 & 34.01 \\
			\hdashline
			NNShot  
			& 55.24\scriptsize $\pm$0.40 & 54.49\scriptsize $\pm$0.91 & 40.21\scriptsize $\pm$1.63 & 49.23\scriptsize $\pm$1.15 & 26.30\scriptsize $\pm$1.21 & 38.91\scriptsize $\pm$0.53 & 24.69\scriptsize $\pm$0.23 & 32.63\scriptsize $\pm$2.59 & 40.21  \\
			StructShot  
			& 53.65\scriptsize $\pm$0.54 & 56.50\scriptsize $\pm$1.17 & 46.86\scriptsize $\pm$0.53 & 53.25\scriptsize $\pm$0.97 & 30.88\scriptsize $\pm$0.96 & 42.80\scriptsize $\pm$0.51 & 27.25\scriptsize $\pm$0.84 & 33.56\scriptsize $\pm$1.06 & 43.10  \\
			\hdashline
			PROTO  
			& 35.78\scriptsize $\pm$0.71 & 47.01\scriptsize $\pm$1.31 & 30.12\scriptsize $\pm$0.77 & 47.13\scriptsize $\pm$0.57 & 15.68\scriptsize $\pm$0.92 & 36.58\scriptsize $\pm$0.87 & 12.68\scriptsize $\pm$0.59 & 28.99\scriptsize $\pm$1.06 & 31.75  \\ 
			CONTaiNER$^\dag$ 
			& 55.95 & 61.83 & 48.35 & 57.12 & 40.43 & 53.70 & 33.84 & 47.49 & 49.84 \\
			ESD$^\dag$
			& 66.46\scriptsize $\pm$0.49 & 74.14\scriptsize $\pm$0.80 & 59.95\scriptsize $\pm$0.69 & 67.91\scriptsize $\pm$1.41 & 41.44\scriptsize $\pm$1.16 & 50.68\scriptsize $\pm$0.94 & 32.29\scriptsize $\pm$1.10 & 42.92\scriptsize $\pm$0.75 & 54.47  \\
			DecomMETA$^\dag$
			& 68.77\scriptsize $\pm$0.24 & 71.62\scriptsize $\pm$0.16 & 63.26\scriptsize $\pm$0.40 & 68.32\scriptsize $\pm$0.10 & 52.04\scriptsize $\pm$0.44 & 63.23\scriptsize $\pm$0.45 & 43.50\scriptsize $\pm$0.59 & 56.84\scriptsize $\pm$0.14 & 60.95  \\
			SpanProto$^\dag$
			& 73.36\scriptsize $\pm$0.18 & 75.19\scriptsize $\pm$0.77 & 66.26\scriptsize $\pm$0.33 & 70.39\scriptsize $\pm$0.63 & 54.49\scriptsize $\pm$0.39 & \bf65.89\scriptsize $\pm$0.82 & 45.39\scriptsize $\pm$0.72 & 59.37\scriptsize $\pm$0.47 & 63.80  \\
			\hline
			\textbf{\met}
			& \bf74.42\scriptsize $\pm$0.61 & \bf76.28\scriptsize $\pm$0.32 & \bf67.91\scriptsize $\pm$0.68 & \bf71.96\scriptsize $\pm$0.35 & \bf55.79\scriptsize $\pm$0.23 & 65.41\scriptsize $\pm$0.35 & \bf47.18\scriptsize $\pm$0.89 & \bf60.71\scriptsize $\pm$0.17 & \bf64.96 \\
            \hline
			\hline
		\end{tabular}
		}
		\caption{F1 scores~(\%) of 5-way 1-shot, 5-way 5-shot, 10-way 1-shot and 10-way 5-shot problems over FEW-NERD dataset. $^\dag$ denotes the results reported in \citet{wang2022spanproto}.
	    We highlight the best results in bold.}
		\label{tab:nerd}
	\end{center}
	
\end{table*}

\begin{table}[htbp]
	\begin{center}
	\resizebox{1.0\columnwidth}{!}{
		\begin{tabular}{ccccc}
			\hline
			\hline
			\multirow{2}{*}{Method}&\multicolumn{4}{c}{\fcm} \\
			\cline{2-5}
            &$5$-$1$&$5$-$5$&$10$-$1$&$10$-$5$ \\
            \hline
			MAML  
			& 28.16\scriptsize $\pm$0.57 & 54.38\scriptsize $\pm$0.37 & 26.23\scriptsize $\pm$0.61 & 44.66\scriptsize $\pm$0.44 \\
			\hdashline
			NNShot  
			& 48.40\scriptsize $\pm$1.27 & 71.55\scriptsize $\pm$1.37 & 41.75\scriptsize $\pm$0.93 & 67.91\scriptsize $\pm$1.51  \\
			StructShot  
			& 48.61\scriptsize $\pm$0.76 & 70.62\scriptsize $\pm$0.83 & 47.77\scriptsize $\pm$0.83 & 65.09\scriptsize $\pm$0.97  \\
			\hdashline
			PROTO  
			& 22.73\scriptsize $\pm$0.86 & 53.95\scriptsize $\pm$0.98 & 22.17\scriptsize $\pm$0.90 & 45.81\scriptsize $\pm$0.99  \\
			CONTaiNER  
			& 57.13\scriptsize $\pm$0.47 & 63.38\scriptsize $\pm$0.68 & 51.87\scriptsize $\pm$0.58 & 60.98\scriptsize $\pm$0.71  \\
			ESD
			& 65.37\scriptsize $\pm$0.79 & 73.29\scriptsize $\pm$0.95 & 58.32\scriptsize $\pm$0.89 & 70.93\scriptsize $\pm$1.01  \\
			DecomMETA
			& 68.01\scriptsize $\pm$0.39 & 72.89\scriptsize $\pm$0.45 & 62.13\scriptsize $\pm$0.28 & 72.14\scriptsize $\pm$0.11  \\
			SpanProto
			& 70.97\scriptsize $\pm$0.41 & 76.59\scriptsize $\pm$0.74 & 63.94\scriptsize $\pm$0.76 & 74.67\scriptsize $\pm$0.33  \\
			\hline
			\textbf{\met}
			& \bf71.89\scriptsize $\pm$0.51 & \bf78.14\scriptsize $\pm$0.36 & \bf65.11\scriptsize $\pm$0.64 & \bf77.58\scriptsize $\pm$0.71  \\
			\hline
			\hline
		\end{tabular}
		}
		\caption{F1 scores~(\%) of 5-way 1-shot, 5-way 5-shot, 10-way 1-shot and 10-way 5-shot problems over \fcm\ dataset. We highlight the best results in bold.}
		\label{tab:comm}
	\end{center}
	
\end{table}

\subsection{Baselines}
\label{baseline}

We compare 
\met\ with eight other few-shot NER models,
which can be grouped into three categories: 
(1) \emph{optimization-based methods}: 
MAML~\cite{MAML_finn2017model}
which adapts to new classes 
by using support instances 
and optimizes the loss of the adapted model 
based on the query instances.
(2) \emph{nearest-neighbor-based methods}: NNShot~\cite{StructShot_yang2020simple} and StructShot~\cite{StructShot_yang2020simple}.
NNShot determines the tag of a query instance based on the word-level distance and
StructShot further improves NNShot by an additional Viterbi decoder.
(3) \emph{prototype-based methods}: PROTO~\cite{PROTO:conf/nips/SnellSZ17}, 
CONTaiNER~\cite{Container_das2021container},
ESD~\cite{ESD:journals/corr/abs-2109-13023},
DecomMETA~\cite{DecomposedMETA_ma2022decomposed}
and
SpanProto~\cite{wang2022spanproto}.
Specifically,
DecomMETA addresses few-shot NER by sequentially tackling few-shot span detection and few-shot entity typing using meta-learning. 
SpanProto transforms the sequential tags into a global boundary matrix and leverage prototypical learning to capture the semantic representations.
For more details of other baselines, see Appendix~\ref{app:baselines}.

\subsection{Experiment Setup}

We implemented \met\ by PyTorch.
The model is initialized by He initialization~\cite{HE:conf/iccv/HeZRS15} and trained by AdamW~\cite{AdamW:journals/corr/abs-1711-05101}.
We run the model for 6,000 epochs with the learning rate 0.2 and the meta learning rate 0.0001 for the improved triplet loss on all the datasets.
For the text encoder, 
we use the pre-trained \texttt{bert-base-Chinese} model for the \fcm\ dataset
and \texttt{bert-base-uncased} model for other datasets.
In the triplet network, we use two feed-forward layers and we set the numbers of hidden units to 1024 and 512.
We also fine-tune the number $T$ of iterations for updating parameters on the support set in each meta-task by grid search over \{1, 3, 5, 7, 9\} 
and set it to 3 on all the datasets.
Moreover,
We set the balancing weight $\alpha$ to 0.3 by grid search over \{0.1, 0.3, 0.5, 0.7, 0.9\}.
For a fair comparison,
we substitute the text encoder as that of \met\ for all the baselines,
use the original codes released by their authors and fine-tune the parameters of the models.
We run all the experiments on a single NVIDIA v100 GPU.
Following~\citet{fewnerd_ding2021few}, 
we evaluate the model performance based on 500 meta-tasks in meta-testing and report the average micro F1-score over 5 runs.
We utilize the \texttt{IO} schema in our experiments, using \texttt{I-type} to denote all the words of a named entity and \texttt{O} to denote other words.
For more details of hyper-parameters, see Appendix~\ref{app:hyper}.
\subsection{Results}

\paragraph{In-domain Experiments}

The results of in-domain experiments in 1-shot and 5-shot settings on FEW-NERD dataset are shown in Table~\ref{tab:nerd}.
From the table,
\met\ consistently outperforms all the baselines on the average F1 score.
For example, 
compared with SpanProto,
\met\ achieves $1.16\%$ improvements on the average F1 score;
when compared against the PROTO model, 
\met\ leads by $33.21\%$ on the average F1 score,
which
clearly demonstrates that our model is very effective in improving PROTO.
On the \fcm\ dataset~(as shown in Table~\ref{tab:comm}),
our model also achieves the best performance across all the settings.
All
these results show that
\met,
which learns adaptive margins
by an improved triplet network,
can perform reasonably well.

\begin{table*}[htbp]
	\begin{center}
	\resizebox{2.08\columnwidth}{!}{
		\begin{tabular}{ccccccc|cc}
			\hline
			\hline
			\multirow{2}{*}{Method}&\multicolumn{2}{c}{WNUT}&\multicolumn{2}{c}{Restaurant}&\multicolumn{2}{c}{Multiwoz}&\multicolumn{2}{c}{Average} \\
			\cline{2-9}
            &$5$-$1$&$5$-$5$&$5$-$1$&$5$-$5$&$5$-$1$&$5$-$5$&$5$-$1$&$5$-$5$ \\
            \hline
			MAML  
			& 17.77\scriptsize $\pm$0.67 & 23.69\scriptsize $\pm$0.71 & 17.53\scriptsize $\pm$0.83 & 22.81\scriptsize $\pm$0.77 & 20.82\scriptsize $\pm$1.01 & 23.61\scriptsize $\pm$0.87 & 18.71 & 23.37  \\
			\hdashline
			NNShot  
			& 15.93\scriptsize $\pm$0.61 & 23.78\scriptsize $\pm$0.67 & 19.37\scriptsize $\pm$0.73 & 32.83\scriptsize $\pm$0.89 & 27.77\scriptsize $\pm$0.91 & 42.19\scriptsize $\pm$1.03 & 21.02 & 32.93  \\
			StructShot  
			& 17.29\scriptsize $\pm$1.01 & 25.18\scriptsize $\pm$0.96 & 20.75\scriptsize $\pm$1.07 & 34.18\scriptsize $\pm$1.18 & 30.79\scriptsize $\pm$1.21 & 44.01\scriptsize $\pm$1.31 & 22.46 & 34.08  \\
			\hdashline
			PROTO  
			& 13.04\scriptsize $\pm$0.71 & 23.20\scriptsize $\pm$0.93 & 15.68\scriptsize $\pm$1.01 & 32.71\scriptsize $\pm$1.07 & 22.09\scriptsize $\pm$0.81 & 41.78\scriptsize $\pm$0.79 & 16.94 & 32.56  \\
			CONTaiNER  
			& 18.15\scriptsize $\pm$1.17 & 19.54\scriptsize $\pm$1.09 & 27.74\scriptsize $\pm$0.89 & 33.41\scriptsize $\pm$0.97 & 34.88\scriptsize $\pm$2.03 & 41.92\scriptsize $\pm$1.93 & 26.92 & 31.62  \\
			ESD
			& 19.24\scriptsize $\pm$0.87 & 26.00\scriptsize $\pm$0.96 & 24.53\scriptsize $\pm$1.03 & 37.85\scriptsize $\pm$0.97 & 35.81\scriptsize $\pm$1.87 & 42.88\scriptsize $\pm$1.05 & 26.53 & 35.58  \\
			DecomMETA
			& 20.98\scriptsize $\pm$0.11 & 31.17\scriptsize $\pm$0.16 & 
			29.75\scriptsize $\pm$0.27 & 41.13\scriptsize $\pm$0.19 &
			33.79\scriptsize $\pm$0.22 & 47.01\scriptsize $\pm$0.36 &
			28.17 & 39.77  \\
            SpanProto
            & 21.94\scriptsize $\pm$0.15 & 32.97\scriptsize $\pm$0.15 & 
			27.75\scriptsize $\pm$0.17 & 39.15\scriptsize $\pm$0.21 &
			36.17\scriptsize $\pm$0.23 & 45.32\scriptsize $\pm$0.35 &
			28.62 & 39.15  \\
            \hline
			\textbf{\met}
			& \bf23.04\scriptsize $\pm$0.78 & \bf34.32\scriptsize $\pm$0.74 & \bf33.01\scriptsize $\pm$0.63 & \bf46.43\scriptsize $\pm$0.57 & \bf41.12\scriptsize $\pm$0.53 & \bf52.73\scriptsize $\pm$0.79 &\bf32.39 & \bf44.49  \\
			\hline
			\hline
		\end{tabular}
		}
    	\caption{F1 scores~(\%) of 5-way 1-shot, 5-way 5-shot problems over three datasets for cross-domain experiments. We highlight the best results in bold.}
    	\label{tab:domain transfer}
	\end{center}
	
\end{table*}

\begin{table*}[htbp]
	\begin{center}
	\resizebox{2.08\columnwidth}{!}{
		\begin{tabular}{ccccccccccccc}
			\hline
			\hline
			\multirow{2}{*}{Method}&\multicolumn{4}{c}{FEW-NERD-INTER}&\multicolumn{4}{c}{FEW-NERD-INTRA}&\multicolumn{4}{c}{\fcm} \\
			\cline{2-13}
            &$5$-$1$&$5$-$5$&$10$-$1$&$10$-$5$&$5$-$1$&$5$-$5$&$10$-$1$&$10$-$5$&$5$-$1$&$5$-$5$&$10$-$1$&$10$-$5$ \\
            \hline
            \met-piw  
			& 64.53 & 65.71 & 55.85 & 56.27 & 44.17 & 53.68 & 33.53 & 43.71
			& 60.46 & 67.69 & 50.75 & 65.47 \\
			\met-piw-rtn  
			& 54.87 & 65.04 & 43.15 & 55.89 & 35.37 & 49.57 & 29.23 & 41.48 
			& 53.13 & 62.89 & 46.72 & 63.09  \\
			\met-otl  
			& 69.73 & 72.91 & 57.70 & 64.31 & 45.28 & 60.21 & 36.61 & 48.56
			& 64.25 & 74.52 & 55.71 & 73.97  \\
			\met-w/o-MAML
			& 72.54 & 73.16 & 65.73 & 70.51 & 53.52 & 61.37 & 43.51 & 54.36 & 70.19 & 72.94 & 61.18 & 74.03   \\
			\hline
			\textbf{\met}
			& \bf74.42 & \bf76.28 & \bf67.91 & \bf71.96 & \bf55.79 & \bf65.41 & \bf47.18 & \bf60.71 & \bf71.89 & \bf78.14 & \bf65.11 & \bf77.58 \\
			\hline
			\hline
		\end{tabular}
		}
		\caption{Ablation study: F1 scores~(\%) of 5-way 1-shot, 5-way 5-shot, 10-way 1-shot and 10-way 5-shot classification over FEW-NERD and \fcm\ datasets. 
	    ``rtn'' means removing triplet network, 
	    ``piw'' means using previous inference way 
	    and ``otl'' means using original triplet loss.
	    We highlight the best results in bold.
	    }
	    \label{tab:ablation}
	\end{center}
	
\end{table*}

\paragraph{Cross-domain Experiments}

We train models on FEW-NERD-INTER~(General) as the source domain and test our models on WNUT~(Social Media), Restaurant~(Review) and Multiwoz~(Dialogue), respectively.
All the three datasets are in different domains from that of FEW-NERD-INTER. 
Since
there is a large generalization gap between the training and test distributions,
cross-domain experiments are generally more challenging than in-domain ones.
Table~\ref{tab:domain transfer}
shows the results.
From the table, 
we see that our model performs very well in both the 1-shot and 5-shot settings.
This clearly shows the generalizability of our model. 

\subsection{Ablation Study}

We conduct an ablation study to understand the characteristics of the main components of \met.
To show the importance of the proposed margin-based inference method,
one variant generates prototype vectors for both entity types and \texttt{O}-class.
In the inference stage,
it computes the distance between a query instance and all these prototype vectors, 
and predict the query instance to be in the class with the smallest distance,
which is similar as previous methods.
We call this variant \textbf{\met-piw} (use \textbf{p}revious \textbf{i}nference \textbf{w}ay).
To study the importance of the triplet network in mapping prototype vectors and samples into a low-dimensional space that is easier to classify,
we further remove the triplet network and replace it with a fully-connected layer. 
Due to the removal of the triplet network, 
adaptive margins cannot be learned,
so we adopt the same inference procedure as in \met-piw.
We call this variant \textbf{\met-piw-rtn} (use \textbf{p}revious \textbf{i}nference \textbf{w}ay and \textbf{r}emove \textbf{t}riplet \textbf{n}etwork ).
To show the importance of the improved triplet loss,
we replace it with the original triplet loss
and call this variant \textbf{\met-otl}\footnote{We fine-tune the margin~$m$ in MeTNet-otl by grid search over \{1, 3, 5, 7, 9\} and set it to 5.}
(\textbf{o}riginal \textbf{t}riplet \textbf{l}oss).
Finally,
we remove the MAML training procedure to explore the impact of MAML on the model
and call this variant \textbf{\met-w/o-MAML}.

The results of ablation study are shown in Table~\ref{tab:ablation}.
From the table,
we observe:
(1)
\met\ beats \met-piw clearly.
For example,
in 5-way 1-shot problem on the \fcm\ dataset, the F1 score of \met\ is $71.89\%$ while that of \met-piw is only $60.46\%$.
This shows that the margin-based inference 
can effectively enhance the model performance.
(2)
The advantage of \met-piw over \met-piw-rtn across all the datasets further shows that the triplet network can learn better embeddings for samples with different classes in the low-dimensional space.
(3)
\met\ leads \met-otl in all the classification tasks.
This  
demonstrates that our improved  triplet loss is highly effective.
(4) Compared against \met-w/o-MAML, 
\met\ leads by $3.55\%$ on the average F1 score, which shows the importance of MAML to the model.

\subsection{Loss Function Analysis}

\begin{figure}[!t]
    \centering
     \includegraphics[scale=0.5]{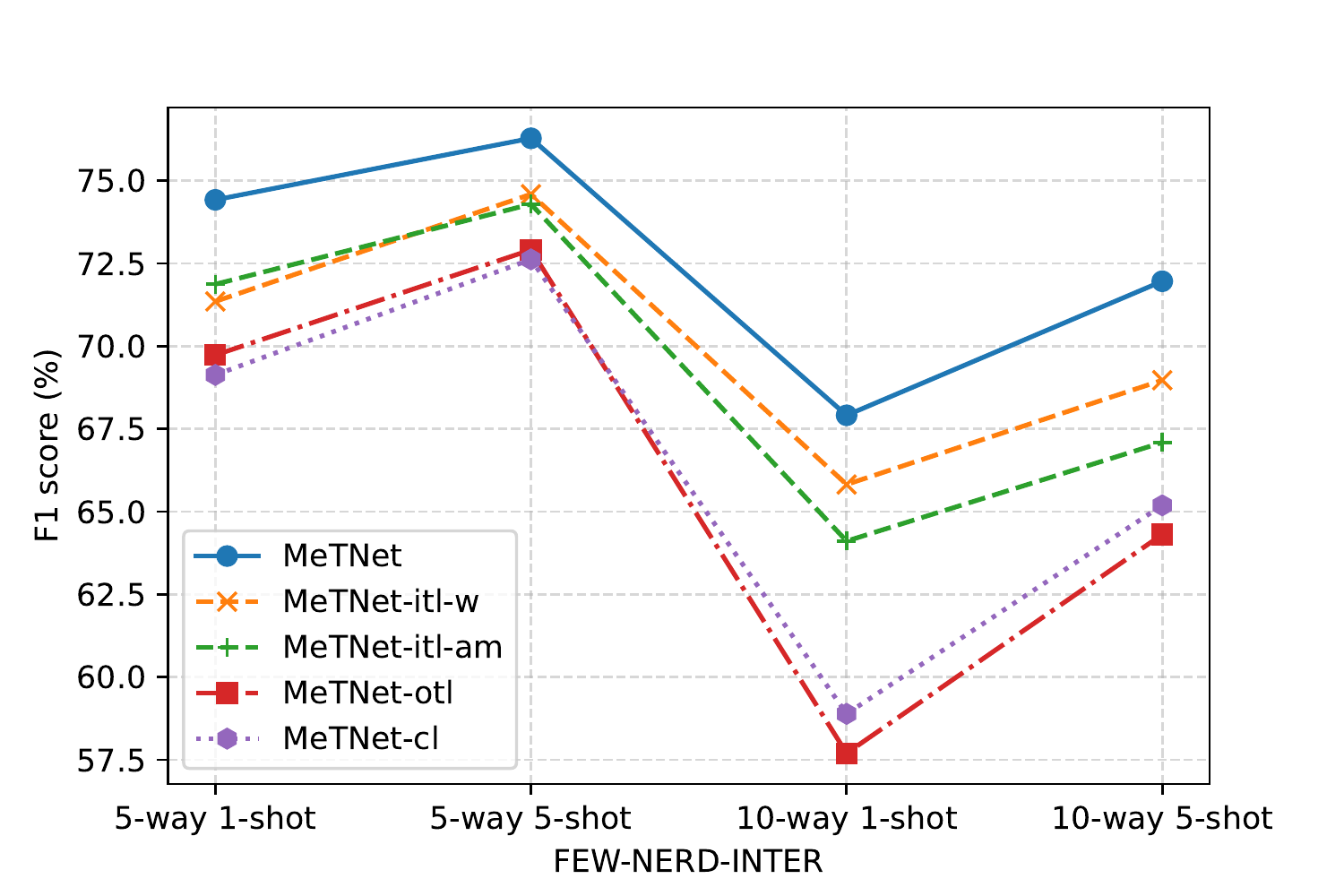}
    \caption{F1 scores~(\%) of 5-way 1-shot, 5-way 5-shot, 10-way 1-shot and 10-way 5-shot classification over FEW-NERD-INTER datasets. 
    ``\textbf{-itl-w}'' means using the \textbf{i}mproved \textbf{t}riplet \textbf{l}oss without important \textbf{w}eights to samples;
    ``\textbf{-itl-am}'' represents using the \textbf{i}mproved \textbf{t}riplet \textbf{l}oss without \textbf{a}daptive \textbf{m}argins (use a fixed margin instead);
    ``\textbf{-otl}''\footnotemark[2] denotes using the \textbf{o}riginal \textbf{t}riplet \textbf{l}oss 
    and ``\textbf{-cl}'' represents using \textbf{c}ontrastive \textbf{l}oss~\cite{contrastive_loss}.}
    \label{fig:loss}
\end{figure}

We next conduct an in-depth experiment 
for loss functions on FEW-NERD-INTER dataset.
The results are shown in Figure~\ref{fig:loss}.
From the results,
we see that MeTNet beats
MeTNet-itl-w clearly,
which demonstrates that 
it is effective that
we assign different weights to samples based on their distances to anchors.
Further,
MeTNet leads MeTNet-itl-am,
which shows that
adaptive margins effectively enhance the model performance.
Moreover,
compared with other loss functions~(e.g. original triplet loss~\cite{tripletNetwork_hoffer2015deep} and contrastive loss~\cite{contrastive_loss}),
we see that MeTNet leads them in all the classification tasks,
which indicates that our improved triplet loss is highly effective.
For other datasets,
we observe similar results that are deferred to Appendix~\ref{app:loss}.

\subsection{Visualization}

\begin{figure}[!t]
    \centering
     \includegraphics[scale=0.366]{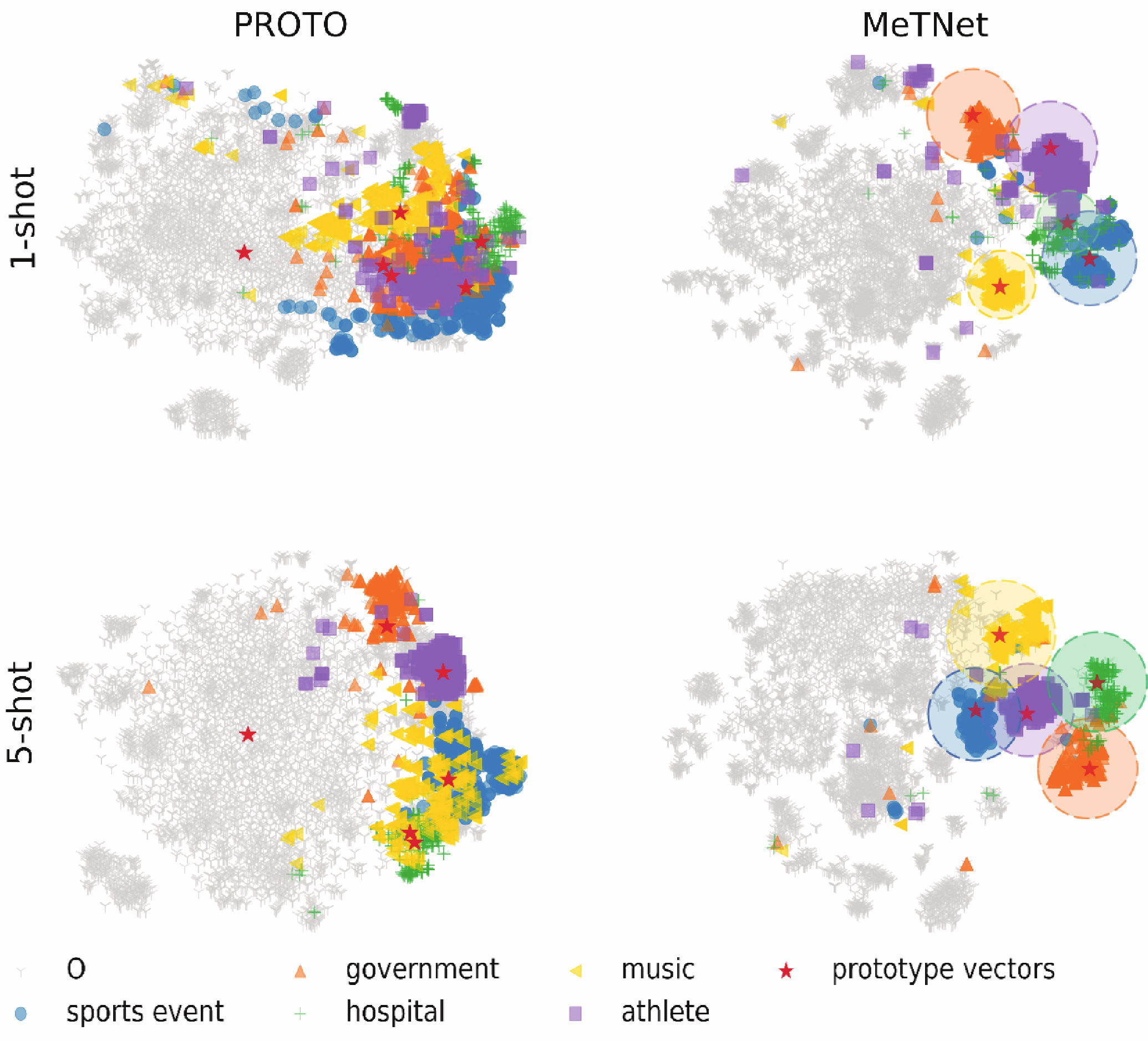}
    \caption{t-SNE visualizations on the FEW-NERD-INTER test sets.
             The representations are obtained from PROTO and \met.
             The dashed circles represent the regions determined by adaptive margins.}
    \label{fig:visual}
\end{figure}

Figure~\ref{fig:visual} visualizes the word-level representations of a query set generated by PROTO and \met\ 
in the 5-way 1-shot and 5-way 5-shot settings on the FEW-NERD-INTER dataset.
Note that PROTO generates prototype vectors for both entity types and \texttt{O}-class,
while \met\ only generates that for entity types.
From the figure,
we see that words in 
\texttt{O}-class are widely distributed,
so using a prototype vector to represent \texttt{O}-class is insufficient.
For those
samples 
closer to other prototype vectors,
they are easily misclassified.
Instead of representing \texttt{O}-class with a prototype vector,
\met\ addresses the problem by learning adaptive margins for entity types only and using a margin-controlled region to make prediction.
Samples outside these regions are labeled with \texttt{O}-class.
Further, 
our method \met\ can generate word embeddings that are clearly separated,
which further explains the effectiveness of \met.

\section{Conclusion}

In this paper,
we studied the few-shot NER problem 
and proposed \met,
which is a meta-learning triplet network with adaptive margins.
As a prototype-based method,
\met\ uses a triplet network to map samples and prototype vectors 
into a low-dimensional space that is easier to be classified.
Further,
to solve the problem that 
\texttt{O}-class is 
semantically complex and thus 
hard to be represented by a prototype vector,
\met\ 
only generates prototype vectors for entity types.
We designed an improved triplet loss function with adaptive margins.
We also presented a margin-based inference procedure to predict the label of a query instance.
We performed extensive experiments in both in-domain and cross-domain settings. 
Experimental results show that 
\met\ can achieve significant performance gains over other state-of-the-art methods.
In particular,
we released the first Chinese few-shot NER dataset~FEW-COMM from a large-scale e-commerce platform,
which aims to provide more insight for future study on few-shot NER.

\section*{Ethics Statement}

The proposed method has no obvious potential risks.
All the scientific artifacts used/created are properly cited/licensed, 
and the usage is consistent with their intended use. 
The paper collects a new dataset~\fcm,  which does not contain any sensitive information.
The dataset is keeping with the rules and reviewed by experts to ensure that it does not create additional risks.
Also, we open up our codes and hyperparameters to facilitate future reproduction without repeated energy cost.


\bibliography{anthology,ref}

\begin{thebibliography}{35}
\expandafter\ifx\csname natexlab\endcsname\relax\def\natexlab#1{#1}\fi

\bibitem[{Budzianowski et~al.(2018)Budzianowski, Wen, Tseng, Casanueva, Ultes,
  Ramadan, and Ga{\v{s}}i{\'c}}]{multiwoz_budzianowski2018multiwoz}
Pawe{\l} Budzianowski, Tsung-Hsien Wen, Bo-Hsiang Tseng, Inigo Casanueva,
  Stefan Ultes, Osman Ramadan, and Milica Ga{\v{s}}i{\'c}. 2018.
\newblock Multiwoz--a large-scale multi-domain wizard-of-oz dataset for
  task-oriented dialogue modelling.
\newblock \emph{arXiv preprint arXiv:1810.00278}.

\bibitem[{Chen et~al.(2021)Chen, Zhang, Li, Xie, Deng, Tan, Huang, Si, and
  Chen}]{LightPrompt_chen2021lightner}
Xiang Chen, Ningyu Zhang, Lei Li, Xin Xie, Shumin Deng, Chuanqi Tan, Fei Huang,
  Luo Si, and Huajun Chen. 2021.
\newblock Lightner: A lightweight generative framework with prompt-guided
  attention for low-resource ner.
\newblock \emph{arXiv preprint arXiv:2109.00720}.

\bibitem[{Cui et~al.(2022)Cui, Hu, Ding, Huang, and
  Liu}]{ProtoPrompt_cui2022prototypical}
Ganqu Cui, Shengding Hu, Ning Ding, Longtao Huang, and Zhiyuan Liu. 2022.
\newblock Prototypical verbalizer for prompt-based few-shot tuning.
\newblock \emph{arXiv preprint arXiv:2203.09770}.

\bibitem[{Cui et~al.(2021)Cui, Wu, Liu, Yang, and
  Zhang}]{BARTNER_cui2021template}
Leyang Cui, Yu~Wu, Jian Liu, Sen Yang, and Yue Zhang. 2021.
\newblock Template-based named entity recognition using bart.
\newblock In \emph{Findings of ACL}, pages 1835--1845.

\bibitem[{Das et~al.(2021)Das, Katiyar, Passonneau, and
  Zhang}]{Container_das2021container}
Sarkar Snigdha~Sarathi Das, Arzoo Katiyar, Rebecca~J Passonneau, and Rui Zhang.
  2021.
\newblock Container: Few-shot named entity recognition via contrastive
  learning.
\newblock \emph{arXiv preprint arXiv:2109.07589}.

\bibitem[{Derczynski et~al.(2017)Derczynski, Nichols, van Erp, and
  Limsopatham}]{wnut_derczynski2017results}
Leon Derczynski, Eric Nichols, Marieke van Erp, and Nut Limsopatham. 2017.
\newblock Results of the wnut2017 shared task on novel and emerging entity
  recognition.
\newblock In \emph{W-NUT}, pages 140--147.

\bibitem[{Devlin et~al.(2018)Devlin, Chang, Lee, and
  Toutanova}]{bert_devlin2018bert}
Jacob Devlin, Ming-Wei Chang, Kenton Lee, and Kristina Toutanova. 2018.
\newblock Bert: Pre-training of deep bidirectional transformers for language
  understanding.
\newblock \emph{arXiv preprint arXiv:1810.04805}.

\bibitem[{Ding et~al.(2021)Ding, Xu, Chen, Wang, Han, Xie, Zheng, and
  Liu}]{fewnerd_ding2021few}
Ning Ding, Guangwei Xu, Yulin Chen, Xiaobin Wang, Xu~Han, Pengjun Xie, Haitao
  Zheng, and Zhiyuan Liu. 2021.
\newblock Few-nerd: A few-shot named entity recognition dataset.
\newblock In \emph{ACL}, pages 3198--3213.

\bibitem[{Finn et~al.(2017)Finn, Abbeel, and Levine}]{MAML_finn2017model}
Chelsea Finn, Pieter Abbeel, and Sergey Levine. 2017.
\newblock Model-agnostic meta-learning for fast adaptation of deep networks.
\newblock In \emph{ICML}, pages 1126--1135.

\bibitem[{Fritzler et~al.(2019)Fritzler, Logacheva, and
  Kretov}]{Proto19:conf/sac/FritzlerLK19}
Alexander Fritzler, Varvara Logacheva, and Maksim Kretov. 2019.
\newblock Few-shot classification in named entity recognition task.
\newblock In \emph{SAC}, pages 993--1000.

\bibitem[{Garnelo et~al.(2018)Garnelo, Rosenbaum, Maddison, Ramalho, Saxton,
  Shanahan, Teh, Rezende, and Eslami}]{CNP_garnelo2018conditional}
Marta Garnelo, Dan Rosenbaum, Christopher Maddison, Tiago Ramalho, David
  Saxton, Murray Shanahan, Yee~Whye Teh, Danilo Rezende, and SM~Ali Eslami.
  2018.
\newblock Conditional neural processes.
\newblock In \emph{ICML}, pages 1704--1713.

\bibitem[{Hadsell et~al.(2006)Hadsell, Chopra, and LeCun}]{contrastive_loss}
Raia Hadsell, Sumit Chopra, and Yann LeCun. 2006.
\newblock Dimensionality reduction by learning an invariant mapping.
\newblock In \emph{CVPR}, volume~2, pages 1735--1742.

\bibitem[{He et~al.(2015)He, Zhang, Ren, and Sun}]{HE:conf/iccv/HeZRS15}
Kaiming He, Xiangyu Zhang, Shaoqing Ren, and Jian Sun. 2015.
\newblock Delving deep into rectifiers: Surpassing human-level performance on
  imagenet classification.
\newblock In \emph{ICCV}, pages 1026--1034.

\bibitem[{Hoffer and Ailon(2015)}]{tripletNetwork_hoffer2015deep}
Elad Hoffer and Nir Ailon. 2015.
\newblock Deep metric learning using triplet network.
\newblock In \emph{SIMBAD}, pages 84--92.

\bibitem[{Hospedales et~al.(2020)Hospedales, Antoniou, Micaelli, and
  Storkey}]{metaSurvey:journals/corr/abs-2004-05439}
Timothy~M. Hospedales, Antreas Antoniou, Paul Micaelli, and Amos~J. Storkey.
  2020.
\newblock Meta-learning in neural networks: {A} survey.
\newblock \emph{CoRR}, abs/2004.05439.

\bibitem[{Hou et~al.(2020)Hou, Che, Lai, Zhou, Liu, Liu, and
  Liu}]{TapNetCDT:conf/acl/HouCLZLLL20}
Yutai Hou, Wanxiang Che, Yongkui Lai, Zhihan Zhou, Yijia Liu, Han Liu, and Ting
  Liu. 2020.
\newblock Few-shot slot tagging with collapsed dependency transfer and
  label-enhanced task-adaptive projection network.
\newblock In \emph{ACL}, pages 1381--1393.

\bibitem[{Huang et~al.(2021)Huang, Li, Subudhi, Jose, Balakrishnan, Chen, Peng,
  Gao, and Han}]{Illi_huang2021few}
Jiaxin Huang, Chunyuan Li, Krishan Subudhi, Damien Jose, Shobana Balakrishnan,
  Weizhu Chen, Baolin Peng, Jianfeng Gao, and Jiawei Han. 2021.
\newblock Few-shot named entity recognition: An empirical baseline study.
\newblock In \emph{EMNLP}, pages 10408--10423.

\bibitem[{Koch et~al.(2015)Koch, Zemel, and Salakhutdinov}]{SN_koch2015siamese}
Gregory Koch, Richard Zemel, and Ruslan Salakhutdinov. 2015.
\newblock Siamese neural networks for one-shot image recognition.
\newblock In \emph{ICML deep learning workshop}, volume~2.

\bibitem[{Liu et~al.(2013)Liu, Pasupat, Cyphers, and
  Glass}]{Restaurant_liu2013asgard}
Jingjing Liu, Panupong Pasupat, Scott Cyphers, and Jim Glass. 2013.
\newblock Asgard: A portable architecture for multilingual dialogue systems.
\newblock In \emph{ICASSP}, pages 8386--8390.

\bibitem[{Loshchilov and Hutter(2017)}]{AdamW:journals/corr/abs-1711-05101}
Ilya Loshchilov and Frank Hutter. 2017.
\newblock Fixing weight decay regularization in adam.
\newblock \emph{CoRR}, abs/1711.05101.

\bibitem[{Ma et~al.(2021)Ma, Zhou, Gui, Tan, Zhang, and
  Huang}]{FudanPrompt_ma2021template}
Ruotian Ma, Xin Zhou, Tao Gui, Yiding Tan, Qi~Zhang, and Xuanjing Huang. 2021.
\newblock Template-free prompt tuning for few-shot ner.
\newblock \emph{arXiv preprint arXiv:2109.13532}.

\bibitem[{Ma et~al.(2022)Ma, Jiang, Wu, Zhao, and
  Lin}]{DecomposedMETA_ma2022decomposed}
Tingting Ma, Huiqiang Jiang, Qianhui Wu, Tiejun Zhao, and Chin-Yew Lin. 2022.
\newblock Decomposed meta-learning for few-shot named entity recognition.
\newblock \emph{arXiv preprint arXiv:2204.05751}.

\bibitem[{Mishra et~al.(2017)Mishra, Rohaninejad, Chen, and
  Abbeel}]{SNAIL_mishra2017simple}
Nikhil Mishra, Mostafa Rohaninejad, Xi~Chen, and Pieter Abbeel. 2017.
\newblock A simple neural attentive meta-learner.
\newblock \emph{arXiv preprint arXiv:1707.03141}.

\bibitem[{Munkhdalai and Yu(2017)}]{Meta_network_munkhdalai2017meta}
Tsendsuren Munkhdalai and Hong Yu. 2017.
\newblock Meta networks.
\newblock In \emph{ICML}, pages 2554--2563.

\bibitem[{Nichol et~al.(2018)Nichol, Achiam, and
  Schulman}]{Reptile_nichol2018first}
Alex Nichol, Joshua Achiam, and John Schulman. 2018.
\newblock On first-order meta-learning algorithms.
\newblock \emph{arXiv preprint arXiv:1803.02999}.

\bibitem[{Ritter et~al.(2012)Ritter, Mausam, Etzioni, and
  Clark}]{IE:conf/kdd/RitterMEC12}
Alan Ritter, Mausam, Oren Etzioni, and Sam Clark. 2012.
\newblock Open domain event extraction from twitter.
\newblock In \emph{KDD}, pages 1104--1112.

\bibitem[{Santoro et~al.(2016)Santoro, Bartunov, Botvinick, Wierstra, and
  Lillicrap}]{MANN_santoro2016meta}
Adam Santoro, Sergey Bartunov, Matthew Botvinick, Daan Wierstra, and Timothy
  Lillicrap. 2016.
\newblock Meta-learning with memory-augmented neural networks.
\newblock In \emph{ICML}, pages 1842--1850.

\bibitem[{Snell et~al.(2017)Snell, Swersky, and
  Zemel}]{PROTO:conf/nips/SnellSZ17}
Jake Snell, Kevin Swersky, and Richard~S. Zemel. 2017.
\newblock Prototypical networks for few-shot learning.
\newblock In \emph{NIPS}, pages 4077--4087.

\bibitem[{Sung et~al.(2018)Sung, Yang, Zhang, Xiang, Torr, and
  Hospedales}]{Relation_network_sung2018learning}
Flood Sung, Yongxin Yang, Li~Zhang, Tao Xiang, Philip~HS Torr, and Timothy~M
  Hospedales. 2018.
\newblock Learning to compare: Relation network for few-shot learning.
\newblock In \emph{CVPR}, pages 1199--1208.

\bibitem[{Tong et~al.(2021)Tong, Wang, Xu, Cao, Liu, Hou, and
  Li}]{MUCO_tong2021learning}
Meihan Tong, Shuai Wang, Bin Xu, Yixin Cao, Minghui Liu, Lei Hou, and Juanzi
  Li. 2021.
\newblock Learning from miscellaneous other-class words for few-shot named
  entity recognition.
\newblock \emph{arXiv preprint arXiv:2106.15167}.

\bibitem[{Vinyals et~al.(2016)Vinyals, Blundell, Lillicrap, Kavukcuoglu, and
  Wierstra}]{vinyals2016matching_network}
Oriol Vinyals, Charles Blundell, Timothy Lillicrap, Koray Kavukcuoglu, and Daan
  Wierstra. 2016.
\newblock Matching networks for one shot learning.
\newblock \emph{arXiv preprint arXiv:1606.04080}.

\bibitem[{Wang et~al.(2022)Wang, Wang, Tan, Qiu, Huang, Huang, and
  Gao}]{wang2022spanproto}
Jianing Wang, Chengyu Wang, Chuanqi Tan, Minghui Qiu, Songfang Huang, Jun
  Huang, and Ming Gao. 2022.
\newblock Spanproto: A two-stage span-based prototypical network for few-shot
  named entity recognition.
\newblock \emph{arXiv preprint arXiv:2210.09049}.

\bibitem[{Wang et~al.(2021)Wang, Xu, Liu, Zhou, Cao, Chang, and
  Sui}]{ESD:journals/corr/abs-2109-13023}
Peiyi Wang, Runxin Xu, Tianyu Liu, Qingyu Zhou, Yunbo Cao, Baobao Chang, and
  Zhifang Sui. 2021.
\newblock An enhanced span-based decomposition method for few-shot sequence
  labeling.
\newblock \emph{CoRR}, abs/2109.13023.

\bibitem[{Wiseman and Stratos(2019)}]{nn_wiseman2019label}
Sam Wiseman and Karl Stratos. 2019.
\newblock Label-agnostic sequence labeling by copying nearest neighbors.
\newblock In \emph{ACL}, pages 5363--5369.

\bibitem[{Yang and Katiyar(2020)}]{StructShot_yang2020simple}
Yi~Yang and Arzoo Katiyar. 2020.
\newblock Simple and effective few-shot named entity recognition with
  structured nearest neighbor learning.
\newblock In \emph{EMNLP}, pages 6365--6375.

\end{thebibliography}
\bibliographystyle{acl_natbib}

\clearpage
\appendix

\section{Pseudocode}
\label{app:alg_MeTNet}
The pseudocode of MeTNet training procedure is summarized in Algorithms~\ref{alg:MeTNet}.

\begin{algorithm}[htbp]
\caption{\met\ Training Procedure}
\label{alg:MeTNet}
\begin{algorithmic}[1]
\small 
  \Require
  Training data $ \{\mathcal{D}_{train}, \mathcal{Y}_{train}\} $;
  $ep$ epochs and the number $T$ of iterations of the model updated by the support set in a task;
  $N$ classes in the support set or the query set;
  $K$ samples in each class in the support set and $L$ samples in each class in the query set;
  the pre-trained BERT parameter $\phi$;
  the model parameter $\theta$;
  the set $\mathcal{M}$ of adaptive margins; 
  \Ensure $\phi$, $\theta$ and $\mathcal{M}$ after training;
  \State Randomly initialize $\theta$ and $\mathcal{M}$;
  \For{each $i \in [1, ep]$}
      \State $\mathcal{Y} \leftarrow \texttt{Sample}(\mathcal{Y}_{train}, N)$;
      \State $\mathcal{S}, \mathcal{Q} \leftarrow \emptyset, \emptyset$;
      \For{$y \in \mathcal{Y}$}
        \State $\mathcal{S} \leftarrow \mathcal{S} \cup \texttt{Sample}(\mathcal{D}_{train}\{y\}, K)$;
        \State $\mathcal{Q} \leftarrow \mathcal{Q} \cup \texttt{Sample}(\mathcal{D}_{train}\{y\}\backslash \mathcal{S}, L)$;
      \EndFor
      \State $\mathcal{H}_{\mathcal{S}}, \mathcal{H}_{\mathcal{Q}} \leftarrow \texttt{BERT}_{\phi}(\mathcal{S}), \texttt{BERT}_{\phi}(\mathcal{Q})$;
      \State $\mathcal{H}_{\mathcal{P}} \leftarrow \emptyset$;
      \For{$y \in \mathcal{Y}$}
        \State $\mathcal{H}_{\mathcal{P}} \leftarrow \mathcal{H}_{\mathcal{P}} \cup \texttt{mean}(\mathcal{H}_{\mathcal{S}}\{y\})$;
      \EndFor
      \For{$t \in T$}
          \State Construct triples by $\mathcal{H}_{\mathcal{S}}, \mathcal{H}_{\mathcal{P}}$;
          \State Input triples to the triplet network;
          \State Calculate $\mathcal{L}_{IT}$ by Equation~\ref{eq:improvedTL};
          \State Update $\theta$ to $\theta'$ by Equation~\ref{eq:maml_support};
      \EndFor
      \State Construct triples by $\mathcal{H}_{\mathcal{Q}}, \mathcal{H}_{\mathcal{P}}$;
      \State Input triples to the triplet network;
      \State Calculate $\mathcal{L}_{IT}$ by Equation~\ref{eq:improvedTL};
      \State Update $\phi$ and $\theta$ based on $\theta'$ by Equation~\ref{eq:maml_query};
  \EndFor
  \State \Return $\phi$, $\theta$ and $\mathcal{M}$
\end{algorithmic}
\end{algorithm}

\section{Statistics of Datasets}
\label{app:datasets}

\begin{table}[htbp]
\begin{center}
\resizebox{1.0\columnwidth}{!}{
\begin{tabular}{ccccc}
\hline
Datasets & \# Sentences & \# Entities & \# Classes & Domain \\
\hline
\fcm & 66.2k & 140.9k & 92 & Commodity \\
FEW-NERD & 188.2k & 491.7k & 66 & General \\
WNUT & 4.7k & 3.1k & 6 & Social Media \\
Restaurant & 9.2k & 15.3k & 8 & Review \\
Multiwoz & 23.0k & 20.7k & 14 & Dialogue \\
\hline
\end{tabular}
}
\caption{
Statistics of datasets.
\# Classes corresponds to the number of pre-defined entity types in a dataset.
}
\label{tab:datasets}
\end{center}
\end{table}

We use four public English datasets
and one new Chinese dataset we proposed.
Statistics of these datasets are given in Table~\ref{tab:datasets}.

\begin{figure*}[!t]
    \centering
    \subfloat[]{
    \label{fig:FewInter}
    \includegraphics[width=0.33\textwidth]{pic/Few-NERD-Inter.pdf}
    }
    \subfloat[]{
    \label{fig:FewIntra}
    \includegraphics[width=0.33\textwidth]{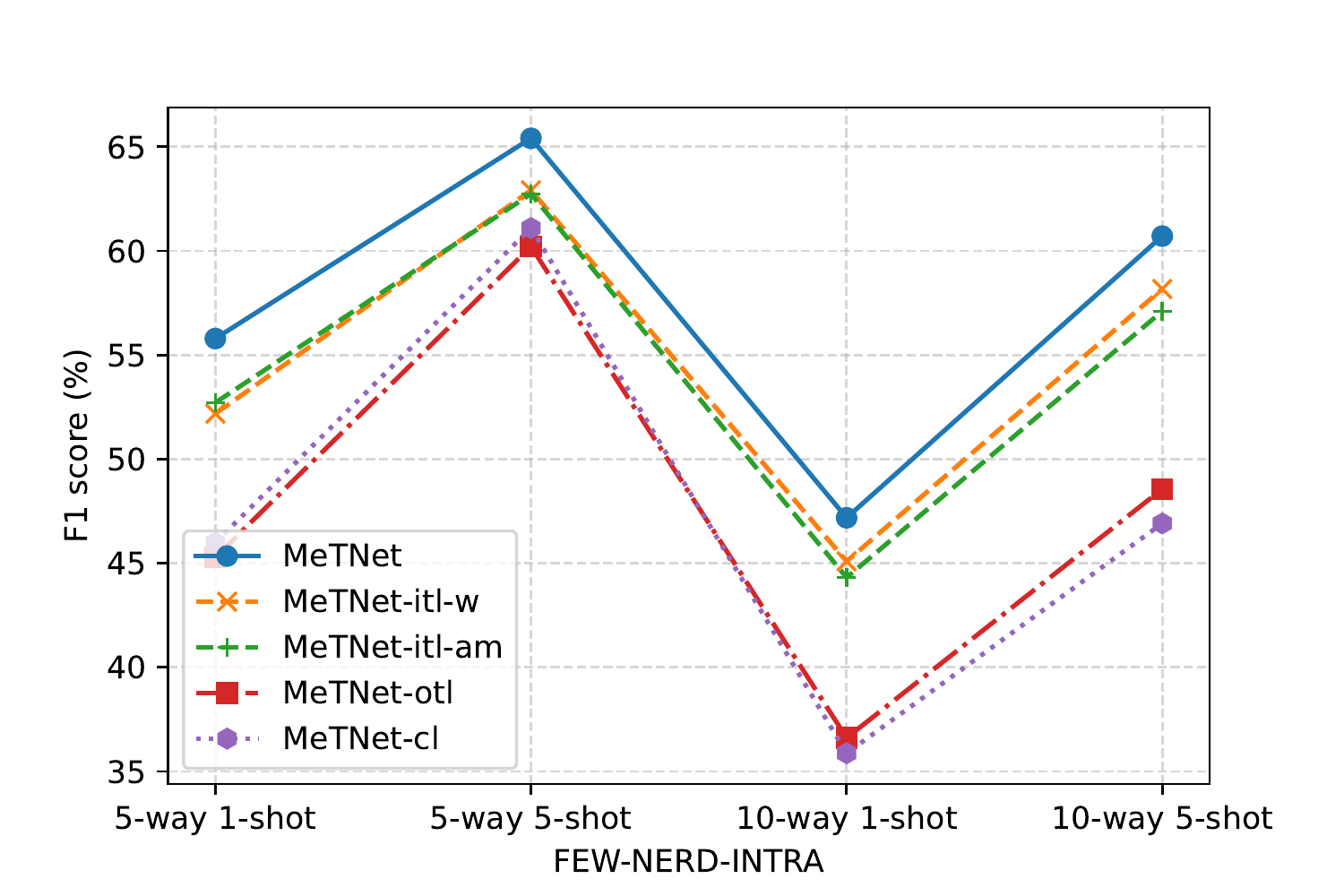}
    }
    \subfloat[]{
    \label{fig:FewComm}
    \includegraphics[width=0.33\textwidth]{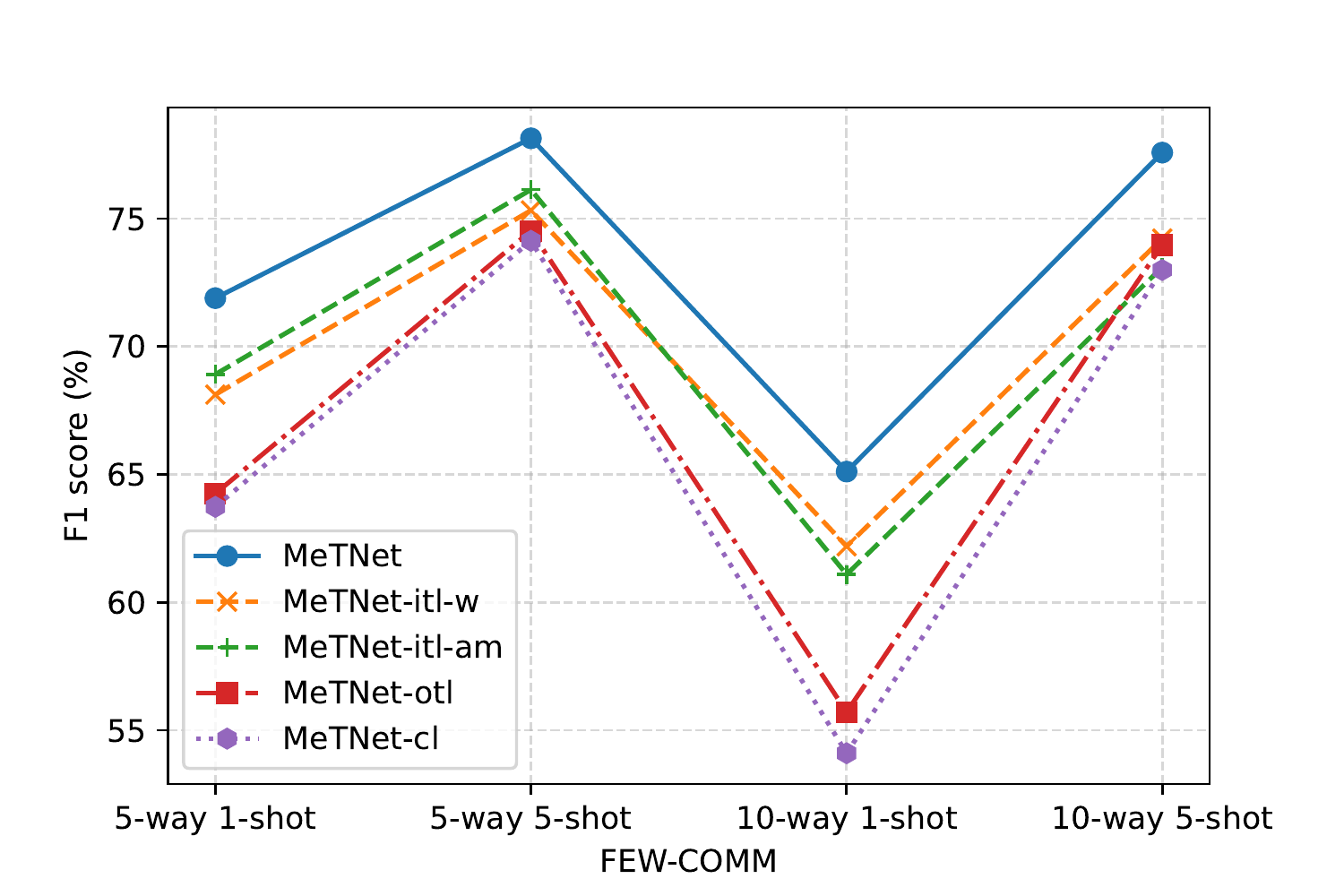}
    }
    \caption{F1 scores~(\%) of 5-way 1-shot, 5-way 5-shot, 10-way 1-shot and 10-way 5-shot classification. 
    ``\textbf{-itl-w}'' means using the \textbf{i}mproved \textbf{t}riplet \textbf{l}oss without important \textbf{w}eights to samples;
    ``\textbf{-itl-am}'' represents using the \textbf{i}mproved \textbf{t}riplet \textbf{l}oss without \textbf{a}daptive \textbf{m}argins (use a fixed margin instead);
    ``\textbf{-otl}'' denotes using the \textbf{o}riginal \textbf{t}riplet \textbf{l}oss 
    and ``\textbf{-cl}'' represents using \textbf{c}ontrastive \textbf{l}oss~\cite{contrastive_loss}.}
    \label{fig:app_loss}
\end{figure*}

\section{\fcm}
\label{appendix:comm}

\subsection{Entity types}
As introduced in Section~\ref{sec:datasets} of the main text, 
\fcm\ is manually annotated with 92 pre-defined entity types, 
and we list all the types and the number of samples belonging to each type in Table~\ref{tab:type}.
We find that since \fcm\ is collected from real application scenarios, 
there is a long-tailed distribution problem, which is a common problem in real scenarios.
How to overcome the influence of long-tailed distribution on the model is a crucial research direction.

\begin{CJK*}{UTF8}{gbsn}
\begin{table*}[htbp]
	\caption{All the pre-defined entity types and the number of samples belonging to each type in \fcm\ dataset.}
	\begin{center}
	\resizebox{2.08\columnwidth}{!}{
		\begin{tabular}{cc|cc|cc|cc}
			\hline
			\hline
			Entity types & \# Samples & Entity types & \# Samples & Entity types & \# Samples & Entity types & \# Samples \\
			\hline
            其他属性 & 44259 & 功能功效 & 13412 & 材质 & 11126 & 适用人群 & 9483 \\
            颜色 & 6955 & 产地 & 4959 & 适用对象 & 2520 & 成分 & 2356 \\
            适用季节 & 1791 & 品质等级 & 1671 & 接口 & 1379 & 适用时间 & 1292 \\
            运输服务 & 1245 & 型号 & 1210 & 商品特色 & 1135 & 国产/进口 & 920 \\
            分类 & 897 & 形状形态 & 874 & 香型 & 860 & 组合形式 & 808 \\
            适用性别 & 801 & 连接方式 & 786 & 控制方式 & 706 & 领型 & 697 \\
            甜度 & 674 & 适用品牌 & 636 & 送礼对象 & 614 & 供电方式 & 585 \\
            面料材质 & 569 & 风味 & 564 & 大小 & 550 & 口感 & 546 \\
            系列 & 530 & 筒高 & 510 & 造型 & 503 & 厚度 & 486 \\
            是否有机 & 483 & 技术类型 & 478 & 厚薄 & 472 & 填充材质 & 469 \\
            适用运营商 & 466 & 袖长 & 465 & 适用车型 & 462 & 糖含量 & 460 \\
            光度 & 457 & 脂肪含量 & 456 & 是否带盖 & 451 & 加热方式 & 447 \\
            长短 & 444 & 版型 & 441 & 适用衣物 & 440 & 资质认证 & 439 \\
            外观 & 436 & 消毒方式 & 430 & 是否清真 & 430 & 部位 & 428 \\
            是否净洗 & 426 & 长度 & 426 & 适用生肖 & 426 & 配件类型 & 424 \\
            袖型 & 422 & 果肉颜色 & 419 & 适用空间 & 419 & 适用燃料 & 416 \\
            适用星座 & 415 & 酸碱度 & 413 & 剂型 & 413 & 锅底类型 & 412 \\
            销售方式 & 412 & 鞋垫材质 & 410 & 适用人数 & 406 & 裙型 & 404 \\
            定制服务 & 403 & 存储容量 & 403 & 成熟状态 & 403 & 是否去皮 & 402 \\
            是否去骨 & 402 & 冲泡方式 & 402 & 赠品 & 401 & 宽度 & 401 \\
            裤长 & 401 & 粗细 & 401 & 礼盒类型 & 400 & 结构 & 400 \\
            色系 & 399 & 净含量 & 376 & 发酵程度 & 321 & 抽数 & 214 \\
            保质期 & 86 & 内容 & 44 & 段位 & 40 & 装订方式 & 11 \\
			\hline
			\hline
		\end{tabular}
		}
		\label{tab:type}
	\end{center}
\end{table*}
\end{CJK*}

\subsection{Splits}
We divided the training set, validation set and test set in a ratio of 6:2:2.
Among them, the training set includes 55 entity types, the validation set includes 18 entity types, and the test set includes 19 entity types.
The entity types contained in the three sets are disjoint.

\subsection{Examples}
We provide some examples on \fcm\ dataset for further understanding, which is shown in Table~\ref{tab:examples}.

\begin{CJK*}{UTF8}{gbsn}

\begin{table*}[htbp]
\renewcommand\arraystretch{2}
	\caption{Examples in \fcm\ dataset. 
	We marked the entities with the corresponding entity types.}
	\begin{center}
	\resizebox{2\columnwidth}{!}{
		\begin{tabular}{c}
			\hline
			\hline
            {\color{red}{日本\small\text{[产地]}}} 
            {\color{orange}{黑色\small\text{[颜色]}}} 
            数字
            {\color{cyan}{帆布\small\text{[材质]}}} 
            烧饼包 
            {\color{cyan}{灯芯绒\small\text{[材质]}}} 
            钱包证件包对开简约 
            {\color{teal}{大容量\small\text{[功能功效]}}} 
            笔袋 \\
            \hline
            {\color{red}{春夏\small\text{[适用季节]}}} 
            爆款{\color{orange}{纯色\small\text{[颜色]}}}
            {\color{cyan}{男女通用\small\text{[适用性别]}}}
            防晒冰袖套
            {\color{violet}{跑男\small\text{[其他属性]}}}
            骑行紫外线护臂【
            {\color{orange}{蓝色\small\text{[颜色]}}} 
            {\color{teal}{直筒\small\text{[版型]}}}  
            无指盒装】 \\
            \hline
            洁丽雅（grace）浴巾a类 
            {\color{red}{纯棉\small\text{[材质]}}} 
            {\color{orange}{加大加厚\small\text{[其他属性]}}}
            {\color{cyan}{成人\small\text{[适用人群]}}} 
            家用柔软 
            {\color{teal}{吸水\small\text{[功能功效]}}}  \\
            \hline
            {\color{red}{精品\small\text{[品质等级]}}} 
            霏慕情趣内衣
            {\color{orange}{女式\small\text{[适用性别]}}}
            性感
            {\color{cyan}{透明\small\text{[颜色]}}}  
            诱惑镂空 
            {\color{teal}{蕾丝\small\text{[材质]}}}
            刺绣
            {\color{violet}{薄纱\small\text{[其他属性]}}}
            7114/2  \\
            \hline
            {\color{red}{金丝绒\small\text{[材质]}}} 
            阔腿裤 
            {\color{orange}{秋冬\small\text{[适用季节]}}}
            {\color{violet}{加绒\small\text{[其他属性]}}}
            高腰垂感宽松
            {\color{cyan}{直筒\small\text{[版型]}}}
            {\color{teal}{显瘦\small\text{[功能功效]}}}
            百搭休闲拖地长裤子  \\
            \hline
            绳子拉车绳 
            {\color{red}{货车\small\text{[适用车型]}}}
            刹车绳子捆绑带拖车绳紧绳器马扎
            {\color{orange}{耐磨\small\text{[功能功效]}}}
            {\color{cyan}{尼龙\small\text{[材质]}}}
            扁带拉紧加粗20米 \\
            \hline
            情趣丝袜修腿
            {\color{red}{显瘦\small\text{[功能功效]}}}
            {\color{orange}{蕾丝\small\text{[材质]}}}
            花边 
            白色
            长筒丝袜
            {\color{cyan}{高筒\small\text{[筒高]}}} 
            情趣连体袜子 \\
            \hline
            电动电瓶车头盔 
            {\color{red}{灰\small\text{[颜色]}}}
            {\color{orange}{女士\small\text{[适用性别]}}}
            {\color{cyan}{夏季\small\text{[适用季节]}}}
            半盔防晒全盔可爱夏天
            {\color{violet}{轻便\small\text{[其他属性]}}}
            安全帽/个  \\
            \hline
            棉拖鞋
            {\color{red}{女士\small\text{[适用性别]}}}
            家居室内厚底防滑月子鞋
            {\color{orange}{冬季\small\text{[适用季节]}}}
            {\color{cyan}{毛绒\small\text{[材质]}}}
            保暖 
            {\color{teal}{情侣\small\text{[适用人群]}}}
            棉鞋
            {\color{violet}{红色\small\text{[颜色]}}}   \\
            \hline
            时尚布艺围裙厨房
            {\color{red}{无袖\small\text{[袖长]}}}
            口袋围腰
            {\color{orange}{成人\small\text{[适用人群]}}}
            格子围裙  \\
            \hline
            【心中最爱】-33朵玫瑰爱心 
            {\color{red}{礼盒\small\text{[形状形态]}}}
            鲜花-送
            {\color{orange}{爱人\small\text{[送礼对象]}}} 
            花店 
            {\color{cyan}{送花上门\small\text{[运输服务]}}}  \\
            \hline
            泳帽
            {\color{red}{女士\small\text{[适用性别]}}}
            长发
            {\color{orange}{防水\small\text{[功能功效]}}}  
            护耳游泳
            {\color{cyan}{硅胶\small\text{[材质]}}} 
            布帽 
            {\color{violet}{舒适\small\text{[其他属性]}}} 
            不勒头帽子游泳泡温泉1个  \\
            \hline
            airism宽松
            {\color{red}{圆领\small\text{[领型]}}}
            t恤
            ( 
            {\color{orange}{五分袖\small\text{[袖长]}}}
            {\color{cyan}{黑色\small\text{[颜色]}}} 
            )  \\
            \hline
            中啡 
            {\color{red}{冷萃\small\text{[冲泡方式]}}} 
            速溶即溶
            {\color{orange}{纯黑\small\text{[颜色]}}} 
            小罐胶囊2gx16颗/盒  \\
			\hline
			\hline
		\end{tabular}
		}
		\label{tab:examples}
	\end{center}
\end{table*}
\end{CJK*}

\section{Baselines}
\label{app:baselines}

We compare 
\met\ with eight other few-shot NER models.

\begin{itemize}

\item
\textbf{MAML}~\cite{MAML_finn2017model}
adapts to new classes 
by using support instances 
and optimizes the loss of the adapted model 
based on the query instances.

\item
\textbf{NNShot}~\cite{StructShot_yang2020simple}
determines the tag of a query instance based on the word-level distance.

\item
\textbf{StructShot}~\cite{StructShot_yang2020simple}
further improves NNShot by an additional Viterbi decoder.

\item
\textbf{PROTO}~\cite{PROTO:conf/nips/SnellSZ17}
computes the prototype vector by averaging all the sample embeddings in the support set for each class.

\item
\textbf{CONTaiNER}~\cite{Container_das2021container}
proposes a contrastive learning method that optimizes the inter-token distribution distance for few-shot NER.

\item
\textbf{ESD}~\cite{ESD:journals/corr/abs-2109-13023}
uses various types of attention based on PROTO to improve the model performance.

\item
\textbf{DecomMETA}~\cite{DecomposedMETA_ma2022decomposed}
addresses few-shot NER by sequentially tackling few-shot span detection and few-shot entity typing using meta-learning.

\item
\textbf{SpanProto}~\cite{wang2022spanproto}
transforms the sequential tags into a global boundary matrix and leverage prototypical learning to capture the semantic representations.

\end{itemize}

\section{Details of Hyper-parameters}
\label{app:hyper}

\begin{table}[htbp]
	\begin{center}
	\resizebox{1.0\columnwidth}{!}{
		\begin{tabular}{cc}
			\hline
			\textbf{Hyper-parameters} & \textbf{Scope}\\
            \hline
			Meta Learning Rate~$\beta$ & \{$1e-5, \textbf{1e-4}, 1e-3, 1e-2, 1e-1$\} \\
			Learning Rate~$\gamma$ & \{$0.1, \textbf{0.2}, 0.3, 0.4, 0.5$\} \\
			Iterations on the support set~$T$ & \{$1, \textbf{3}, 5, 7, 9 $\} \\
			Balancing Weight~$\alpha$ & \{$0.1, \textbf{0.3}, 0.5, 0.7, 0.9$\} \\
			Dropout Rate~$\epsilon$ & \{\textbf{0.1}, 0.2, 0.3\} \\
			Batch Size~$BS$ & \{\textbf{1}, 2, 3, 4, 5\} \\
			\hline
		\end{tabular}
		}
		\caption{The searching scope for all hyper-parameters. We highlight the best settings in bold. Note that the batch size in the $N$-way $K$-shot setting represents the number of episodes in one batch.}
		\label{tab:hyper}
	\end{center}
	
\end{table}

The searching scope of each hyper-parameter is shown in Table~\ref{tab:hyper}.
The model is initialized by He initialization~\cite{HE:conf/iccv/HeZRS15} and trained by AdamW~\cite{AdamW:journals/corr/abs-1711-05101}.
We run the model for 6,000 epochs with the learning rate 0.2 and the meta learning rate 0.0001 for the improved triplet loss on all the datasets.
For the text encoder, 
we use the pre-trained \texttt{bert-base-Chinese} model for the \fcm\ dataset
and \texttt{bert-base-uncased} model for other datasets.
In the triplet network, we use two feed-forward layers and we set the numbers of hidden units to 1024 and 512.
We also fine-tune the number $T$ of iterations for updating parameters on the support set in each meta-task by grid search over \{1, 3, 5, 7, 9\} 
and set it to 3 on all the datasets.
Moreover,
We set the balancing weight $\alpha$ to 0.3 by grid search over \{0.1, 0.3, 0.5, 0.7, 0.9\}.
We run all the experiments on a single NVIDIA v100 GPU.
Following~\citet{fewnerd_ding2021few}, 
we evaluate the model performance based on 500 meta-tasks in meta-testing and report the average micro F1-score over 5 runs.
We utilize the \texttt{IO} schema in our experiments, using \texttt{I-type} to denote all the words of a named entity and \texttt{O} to denote other words.

\section{Loss Function Analysis}
\label{app:loss}

We conduct an in-depth experiment 
for the loss function.
The results are shown in Figure~\ref{fig:app_loss}.
From the results,
we see that MeTNet beats
MeTNet-itl-w and MeTNet-itl-am clearly,
which demonstrates that
the our improvements including sample weights and adaptive margins effectively enhance the model performance.
Further,
compared with other loss functions~(e.g. triplet loss~\cite{tripletNetwork_hoffer2015deep} and contrastive loss~\cite{contrastive_loss}),
we see that MeTNet leads them in all the classification tasks,
which demonstrates that our improved triplet loss is highly effective.

\end{document}